\documentclass[sigconf]{acmart}
\AtBeginDocument{%
  }

\copyrightyear{2026}
\acmYear{2026}
\setcopyright{cc}
\setcctype{by}
\acmConference[KDD '26]{Proceedings of the 32nd ACM SIGKDD Conference on Knowledge Discovery and Data Mining V.2}{August 09--13, 2026}{Jeju Island, Republic of Korea}
\acmBooktitle{Proceedings of the 32nd ACM SIGKDD Conference on Knowledge Discovery and Data Mining V.2 (KDD '26), August 09--13, 2026, Jeju Island, Republic of Korea}
\acmDOI{10.1145/3770855.3817561}
\acmISBN{979-8-4007-2259-2/2026/08}

\usepackage{booktabs}
\usepackage{tabularx}
\usepackage{multirow}
\usepackage{graphicx}
\usepackage{subcaption}
\usepackage{xcolor}
\newcommand{\cn}[1]{\begin{CJK*}{UTF8}{gbsn}#1\end{CJK*}}
\usepackage[table]{xcolor}
\usepackage{colortbl}
\usepackage{xfp}
\usepackage{tcolorbox}
\usepackage{float}
\usepackage{CJKutf8}

\definecolor{acaBlue}{RGB}{60, 100, 160}
\newcommand{\cc}[1]{%
    \ifnum\fpeval{#1>65}=1%
        \textcolor{white}{\cellcolor{acaBlue!#1}#1}%
    \else%
        \cellcolor{acaBlue!#1}#1%
    \fi%
}

\definecolor{acadRedStrong}{HTML}{F4BBD0}
\definecolor{acadRedMed}{HTML}{F8D3DF}
\definecolor{acadRedLight}{HTML}{FCEEF3}

\definecolor{acadGreenLight}{HTML}{EBF5E9}
\definecolor{acadGreenMed}{HTML}{C8E6C9}

\newcommand{\ccRR}[1]{\cellcolor{acadRedStrong}#1}
\newcommand{\ccR}[1]{\cellcolor{acadRedMed}#1}
\newcommand{\ccr}[1]{\cellcolor{acadRedLight}#1}

\newcommand{\ccN}[1]{#1}

\newcommand{\ccg}[1]{\cellcolor{acadGreenLight}#1}
\newcommand{\ccG}[1]{\cellcolor{acadGreenMed}#1}

\definecolor{captionRed}{HTML}{C62828}  
\definecolor{captionGreen}{HTML}{2E7D32} 

\begin{document}

\title{PolySpeech-100: A Large-Scale Benchmark for Speech Understanding Across 100+ Languages and Dialects}

\author{Sicheng Yang}
\email{yangsc25@mails.tsinghua.edu.cn}
\author{Shulan Ruan}
\email{slruan@sz.tsinghua.edu.cn}
\author{Shiwei Wu}
\email{davidwu16@sz.tsinghua.edu.cn}
\affiliation{%
  \institution{Shenzhen International Graduate School,\\ Tsinghua University}
  \city{Shenzhen}
  \country{China}
}
\author{Yu Liu}
\authornote{Corresponding author.}
\email{liuyu_thu@mail.tsinghua.edu.cn}
\affiliation{%
  \institution{Department of Electronic Engineering,\\ Tsinghua University}
  \city{Beijing}
  \country{China}
}

\author{Lu Fan}
\email{fanlu@jd.com}
\affiliation{%
  \institution{JD AI Research}
  \city{Beijing}
  \country{China}
}

\author{Zhi Li}
\email{zhilizl@sz.tsinghua.edu.cn}
\author{You He}
\email{heyou@mail.tsinghua.edu.cn}
\affiliation{%
  \institution{Shenzhen International Graduate School,\\ Tsinghua University}
  \city{Shenzhen}
  \country{China}
}

\renewcommand{\shortauthors}{Sicheng Yang et al.}


\settopmatter{printacmref=true} 

\begin{abstract}

While End-to-End (E2E) Speech-Large Language Models (Speech-LLMs) are rapidly evolving, their evaluation methodologies remain limited to the era of simple transcription. Existing benchmarks suffer from three critical limitations: a pronounced bias towards high-resource languages, a focus on low-level recognition (ASR) rather than semantic reasoning, and a neglect of regional dialects. To bridge this gap, we introduce PolySpeech-100, a massive-scale benchmark designed to assess `native-level' speech comprehension across 110 linguistic variants.
We employ a novel hybrid construction pipeline that augments gold-standard human recordings with instruction-driven synthetic speech, allowing us to cover 19 distinct Chinese dialects and over 80 low-resource languages. Extensive evaluation of 22 state-of-the-art models (including Gemini-3, GPT-Audio, and Qwen2.5-Omni) yields pivotal insights. First, we demonstrate that open-source E2E models outperform Cascade (ASR+LLM) systems on heavy dialects, proving that direct audio processing preserves critical paralinguistic cues and prosodic features (e.g., intonation, stress) that are often lost in standard transcription. Second, we reveal a significant performance gap: while commercial models maintain robustness, open-source models suffer catastrophic degradation on low-resource languages. 
Finally, counter-intuitively, we observe that under standard zero-shot settings, Chain-of-Thought prompting frequently degrades speech understanding performance for most evaluated models, revealing a potential modality alignment gap in current architectures.
PolySpeech-100 establishes a rigorous standard for the next generation of inclusive, omni-capable Speech-LLMs. 
The data, demo, and code are publicly available at \url{https://github.com/YoungSeng/PolySpeech-100}.

\end{abstract}

%


\begin{teaserfigure}
  \includegraphics[width=\textwidth]{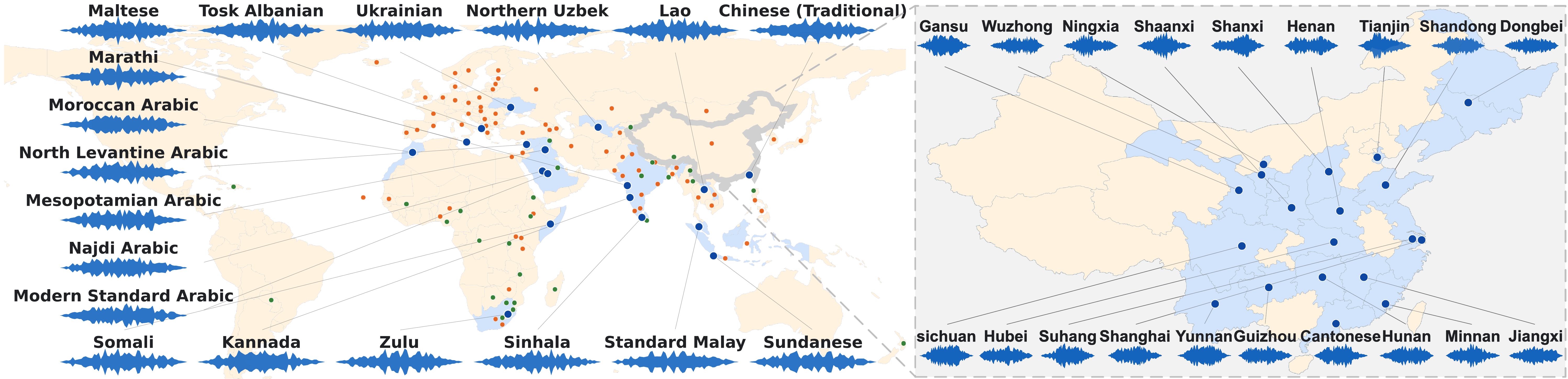}
  \caption{Geographic distribution of PolySpeech-100. The points represent specific languages and dialects included in the benchmark. Orange shading indicates coverage of mainstream languages, while blue shading highlights regions covered by our newly added languages. The right panel highlights the common Chinese dialects included in the benchmark.}
  \Description{Enjoying the baseball game from the third-base
  seats. Ichiro Suzuki preparing to bat.}
  \label{fig:teaser}
\end{teaserfigure}

\maketitle


\section{Introduction}

The field of Artificial Intelligence is undergoing a rapid transition from text-based Large Language Models (LLMs) to End-to-End (E2E) Speech-LLMs \cite{wang2025towards, DBLP:journals/corr/abs-2411-13577}. Recent systems like ChatGPT \cite{DBLP:journals/corr/abs-2410-21276} and Gemini \cite{google2025gemini3flash} have demonstrated remarkable capabilities in perceiving and understanding directly with speech \cite{DBLP:conf/acl/YangSYC25}. Unlike traditional cascaded systems, which first transcribe speech to text (ASR) and then process the text, these native multimodal models aim to capture paralinguistic cues, emotional tone, and dialectal nuances that are often lost in transcription \cite{DBLP:journals/corr/abs-2504-08528, DBLP:journals/corr/abs-2509-22727, DBLP:conf/acl/Chen0YLLXN00L0025}.

Despite this progress, the evaluation infrastructure for Speech-LLMs lags significantly behind model development \cite{DBLP:conf/acl/CuiYJMZWGK25, DBLP:journals/corr/abs-2402-13236}. 
Existing benchmarks suffer from three primary limitations. 
First, they are heavily \textbf{skewed towards high-resource languages}, predominantly English and Standard Mandarin, leaving the vast majority of the world's languages untested. 
Second, most benchmarks focus on \textbf{lower-level tasks} like Automatic Speech Recognition (ASR) or simple instruction following, rather than complex reasoning or semantic understanding. 
Third, and most critically, there is a lack of diverse \textbf{dialectal evaluation}. Current datasets often treat languages as monoliths (e.g., `Chinese' or `Arabic'), ignoring the rich regional variations—such as Cantonese, Sichuanese, or Maghrebi Arabic—that pose the greatest challenges to real-world deployment.

We introduce PolySpeech-100, a benchmark designed to shift speech evaluation from transcription (`hearing') to \textbf{content comprehension (`understanding') across 100+ languages and dialects}. Built upon the Belebele dataset \cite{DBLP:conf/acl/BandarkarLMASHG24, DBLP:conf/acl/Costa-jussaYAAC25}, this high-fidelity corpus challenges models to perform spoken question answering rather than relying on simple pattern matching. Notably, PolySpeech-100 emphasizes fine-grained dialectal diversity by incorporating 19 distinct Chinese regional variants (e.g., Minnan), allowing for a rigorous assessment of whether models possess true `native-level' listening capabilities or merely overfit to standard broadcast speech.

To support this scale, we propose a hybrid data construction pipeline that effectively solves data scarcity issues for low-resource languages. 
Our human verification ($r=0.83$ correlation) proves that high-quality synthesis is a valid proxy for evaluating dialectal robustness when human data is scarce.
To the best of our knowledge, PolySpeech-100 represents the most linguistically diverse speech comprehension benchmark to date. It provides a comprehensive evaluation of current speech-centric models, establishing a new standard for the community to measure progress toward inclusive speech understanding.
We evaluated 22 state-of-the-art models, including proprietary APIs (e.g., Gemini \cite{google2025gemini3flash}) and open-source E2E models (e.g., MiMo-Audio \cite{DBLP:journals/corr/abs-2512-23808}, Step-Audio-2 \cite{DBLP:journals/corr/abs-2507-16632}), as well as traditional ASR+LLM pipelines. Our experiments yield several critical insights:
(1) For heavy dialects, open-source E2E models significantly outperform traditional ASR pipelines, acoustic features essential for semantic disambiguation that ASR systems typically filter out.
(2) As for low-resource languages (e.g., Zulu, Lao), most open-source models degrade significantly while commercial models maintain robustness, highlighting a critical direction for future research.
(3) 
In terms of reasoning strategies, we find that under our evaluated settings, many current Speech-LLMs struggle with Chain-of-Thought (CoT) in the audio modality, often performing better with direct answers than with intermediate reasoning steps.

\section{Related Work}
\subsection{Traditional ASR and Translation Datasets}
Early research focused on Automatic Speech Recognition (ASR) and Speech-to-Text Translation (S2TT). 
For English and multilingual ASR, LibriSpeech \cite{DBLP:conf/icassp/PanayotovCPK15}, GigaSpeech \cite{DBLP:conf/interspeech/ChenCWDZWSPTZJK21}, and Multilingual LibriSpeech (MLS) \cite{DBLP:conf/interspeech/PratapXSSC20} remain the standard.
For Chinese, datasets like AISHELL \cite{DBLP:journals/corr/abs-1709-05522, DBLP:journals/corr/abs-1808-10583}, WenetSpeech \cite{DBLP:conf/icassp/ZhangLGSYXXBCZW22}, and KeSpeech \cite{DBLP:conf/nips/Tang0XSLZWTXZYL21} cover Mandarin in diverse scenarios.
Code-switching scenarios are specifically addressed in benchmarks like Mandarin-English
Code-Switching Challenge dataset \cite{DBLP:journals/corr/abs-2007-05916}.
To evaluate multilingual capabilities, researchers use Common Voice \cite{DBLP:conf/lrec/ArdilaBDKMHMSTW20}, which includes many languages but focuses on short sentence reading. 
Fleurs \cite{DBLP:conf/slt/ConneauMKZADRRB22} provides n-way parallel speech data for 102 languages, used for both ASR and translation tasks. 
Similarly, CoVoST 2 \cite{DBLP:conf/interspeech/WangWGP21} offers large-scale speech translation benchmarks.
Voxpopuli \cite{DBLP:conf/acl/WangRLWTHWPD20} provides European language data from political speeches.
However, these datasets mainly evaluate transcription quality (WER or BLEU scores). They do not measure the semantic understanding or reasoning capabilities required by modern Large Language Models (LLMs).

\subsection{Speech Understanding and Paralinguistics}
Beyond transcription, speech contains rich information regarding emotion, environment, and speaker intent.
Datasets like MELD \cite{DBLP:conf/acl/PoriaHMNCM19} and IEMOCAP \cite{DBLP:journals/lre/BussoBLKMKCLN08} focus on emotion recognition in dialogue.
VocalSound \cite{DBLP:conf/icassp/GongYG22} evaluates the classification of non-speech human sounds (e.g., laughter, sighing).
In the general audio-text domain, WavCaps \cite{DBLP:journals/taslp/MeiMLKKZPZW24}, Clotho \cite{DBLP:conf/icassp/DrossosLV20}, AudioCaps \cite{DBLP:conf/naacl/KimKLK19} and MusicCaps \cite{DBLP:conf/ismir/DohCLN23} test captioning capabilities for environmental sounds and music.
More complex reasoning and environmental understanding are evaluated in MMAR \cite{DBLP:journals/corr/abs-2505-13032}, OmniBench \cite{DBLP:journals/corr/abs-2409-15272}, WorldSense \cite{DBLP:journals/corr/abs-2502-04326}, and DailyOmni \cite{DBLP:journals/corr/abs-2505-17862}.
For dialogue understanding, the Fisher \cite{cieri2004fisher} dataset provides telephone conversation data, which is often used to test reference resolution and topic modeling.
SpokenWOZ \cite{DBLP:conf/nips/SiMGWLD0Y0L23}, MultiChallenge Audio \cite{DBLP:conf/acl/DeshpandeSMJHLK25} and SLURP \cite{DBLP:conf/emnlp/BastianelliVSR20} are designed for dialogue state tracking and spoken language understanding (SLU), providing multi-turn conversation data.
While valuable, these tasks are often specific classifications (e.g., `happy' vs. `sad') or short-context descriptions rather than chat or reasoning across diverse languages.

\begin{figure*}[h!]
  \centering
  \includegraphics[width=0.7125\linewidth]{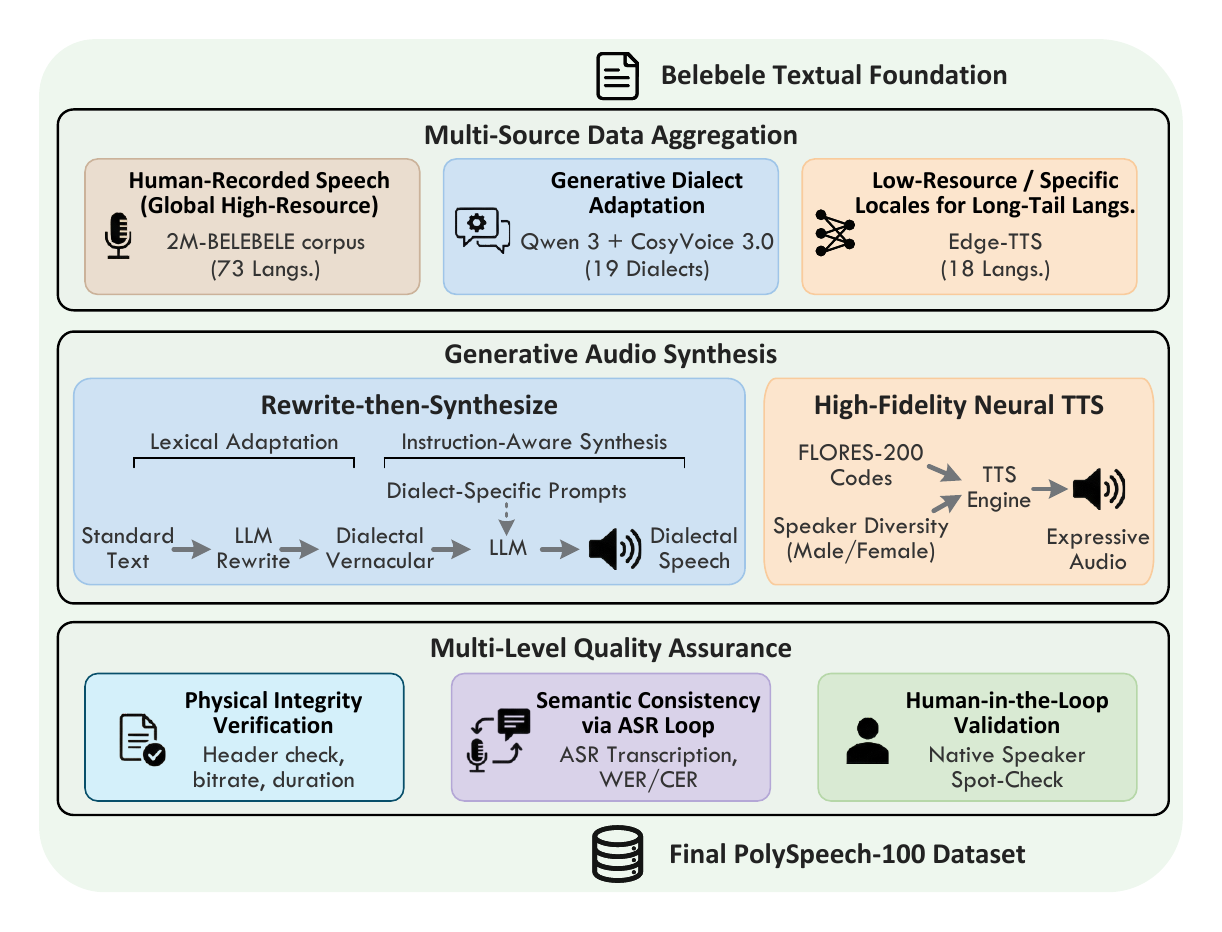}
  \caption{The data construction pipeline of PolySpeech-100. The framework consists of three stages: multi-source data aggregation, generative audio synthesis via a Rewrite-then-Synthesize strategy, and multi-level quality assurance.}
  \label{fig:1}
  
\end{figure*}

\subsection{Benchmarks for Speech-LLM}    
With the rise of Speech-LLMs, comprehensive benchmarks have emerged.
AIR-Bench \cite{DBLP:conf/acl/YangXLC0ZLLZZZ24} is a comprehensive benchmark covering speech, sound, and music. It uses datasets like Fisher \cite{cieri2004fisher} and SpokenWOZ \cite{DBLP:conf/nips/SiMGWLD0Y0L23}.
Dynamic-SUPERB \cite{DBLP:conf/icassp/HuangLWHKWACSPS24} evaluates instruction-following abilities across various tasks.
Recent suites like OpenAudioBench \footnote{\url{https://huggingface.co/datasets/baichuan-inc/OpenAudioBench}}, UltraEval-Audio \cite{shi2026ultraeval}, URO-Bench \cite{DBLP:journals/corr/abs-2502-17810}, and Big Bench Audio \footnote{\url{https://huggingface.co/datasets/ArtificialAnalysis/big_bench_audio}} evaluate instruction following and reasoning.
Instruction-tuning evaluations include InstructS2S \cite{DBLP:conf/iclr/FangGZMZ025}, Moss \cite{DBLP:journals/ijautcomp/SunZHLCLYSTZZCZZLZZLYWY24}, and Alpaca-Eval \cite{alpaca_eval}.
VoiceBench \cite{DBLP:journals/corr/abs-2410-17196} focuses on safety, instruction following, and reasoning (QA). It adapts text benchmarks (e.g., AlpacaEval) into speech.
MMAU \cite{DBLP:conf/iclr/SakshiTKSSNDGM25} and MMSU \cite{DBLP:journals/corr/abs-2506-04779} test multi-task understanding and reasoning, including math and logic in audio.

Many recent models propose their own internal evaluation sets.
Qwen2-Audio \cite{DBLP:journals/corr/abs-2407-10759}, Qwen2.5/3-Omni \cite{DBLP:journals/corr/abs-2503-20215, DBLP:journals/corr/abs-2509-17765}, and Qwen3.5-Omni \cite{DBLP:journals/corr/abs-2604-15804} test on massive multi-task collections.
LLaMA-Omni \cite{DBLP:conf/iclr/FangGZMZ025} and LLaMA-Omni 2 \cite{DBLP:journals/corr/abs-2505-02625} focus on low-latency speech interaction, evaluating content and style.
Mini-Omni \cite{DBLP:journals/corr/abs-2408-16725} and Moshi \cite{DBLP:journals/corr/abs-2410-00037} explore real-time streaming capabilities.
SLAM-LLM \cite{ma2026slam} investigates speech encoders with datasets like GigaSpeech 2 \cite{DBLP:conf/acl/0005SZCLYD0LWL025}.
Other notable models include Fun-Audio-Chat \cite{DBLP:journals/corr/abs-2512-20156}, MinMo \cite{DBLP:journals/corr/abs-2501-06282}, Step-Audio 2 \cite{DBLP:journals/corr/abs-2507-16632}, MiMo-Audio \cite{DBLP:journals/corr/abs-2512-23808}, Ming-Omni \cite{DBLP:journals/corr/abs-2506-09344}, Baichuan-Audio \cite{DBLP:journals/corr/abs-2502-17239}, Kimi-Audio \cite{DBLP:journals/corr/abs-2504-18425} and X-Talk \cite{DBLP:journals/corr/abs-2512-18706}.
Most of these works evaluate English and Mandarin well. However, they lack comprehensive testing for low-resource languages and regional dialects (e.g., Sichuan, Cantonese) in a complex reasoning context.

\subsection{Synthetic Data for Evaluation}
Collecting human recordings for rare dialects is expensive and time-consuming \cite{DBLP:journals/corr/abs-2503-20212, DBLP:journals/corr/abs-2408-00284}. 
A growing trend in recent model evaluations (e.g., for LLaMA-Omni2 \cite{DBLP:journals/corr/abs-2505-02625}, Mini-Omni \cite{DBLP:journals/corr/abs-2408-16725}, and MiMo-Audio \cite{DBLP:journals/corr/abs-2512-23808}) is to convert text-based benchmarks \cite{DBLP:conf/iclr/FangGZMZ025, DBLP:journals/ijautcomp/SunZHLCLYSTZZCZZLZZLYWY24} \footnote{\url{https://huggingface.co/datasets/XiaomiMiMo/SpeechMMLU}}\textsuperscript{,}\footnote{\url{https://huggingface.co/collections/mistralai/speech-evals}} into audio using Text-to-Speech (TTS) systems. 
However, these TTS-based evaluations often lack real-world noise and linguistic variations (`clean' data issue).
In contrast, 2M-BELEBELE \cite{DBLP:conf/acl/Costa-jussaYAAC25} serves as a gold standard for multilingual reading comprehension with high-quality human recordings, though its coverage of specific dialectal nuances and code-switching remains limited.
To address these gaps, we propose PolySpeech-100, a large-scale hybrid benchmark covering 100+ languages and dialects. By combining human-recorded gold standards with diverse synthesized data covering dialects and long-tail languages, we provide a robust assessment of Speech-LLMs' `Babel' capabilities in complex, real-world linguistic scenarios.

\section{Dataset Creation}
\label{sec:dataset_creation}

We introduce PolySpeech-100, a scalable and linguistically diverse speech understanding benchmark covering over 100 language variants. To address the scarcity of evaluation data for dialects and low-resource languages \cite{DBLP:conf/acl/BarteldsSMJ023, DBLP:conf/interspeech/Xu00000LL025}, we propose a hybrid construction pipeline that augments human recordings with state-of-the-art synthetic generation. Our methodology follows a three-stage framework: (1) Multi-Source Data Aggregation \cite{DBLP:conf/acl/BarteldsSMJ023,DBLP:conf/interspeech/LeeWLL18}, (2) Generative Audio Synthesis, and (3) Multi-Level Quality Assurance.

\subsection{Data Foundation: The Belebele Backbone} 
To ensure rigorous cross-lingual comparability, we utilize the \textsc{Belebele} benchmark~\cite{DBLP:conf/acl/BandarkarLMASHG24} as our textual foundation. \textsc{Belebele} offers parallel reading comprehension passages and multiple-choice questions (MCQs) aligned across 122 languages. This parallel structure is ideal for speech understanding as it allows for controlled evaluation of acoustic modeling capabilities across diverse language families without semantic variation.

\subsection{Multi-Source Audio Construction} Our dataset construction employs a stratified approach to audio acquisition, balancing authenticity with coverage. We categorize our data sources into three distinct tracks.

\subsubsection{Track 1: Human-Recorded Speech (High-Resource)} 

For the core set of 73 languages, we leverage the 2M-BELEBELE corpus \cite{DBLP:conf/acl/Costa-jussaYAAC25}. This component provides professionally recorded human speech, serving as the gold standard for acoustic naturalness. We parse the source data to extract aligned tuples of (Passage, Question, Options), ensuring that high-resource languages (e.g., English, Spanish, French) are represented by authentic native speaker recordings.

\subsubsection{Track 2: Generative Dialect Adaptation.} 
A critical limitation of existing speech benchmarks is the neglect of regional dialects. To bridge this gap,  we propose a high-fidelity generation pipeline that challenges the view that synthetic data is unsuitable for evaluation~\cite{DBLP:conf/acl/Costa-jussaYAAC25}. 
While prior studies relied on concatenation-based systems, we utilize CosyVoice 3.0~\cite{DBLP:journals/corr/abs-2505-17589}, a state-of-the-art generative speech model with zero-shot instruction following.
Unlike traditional TTS \cite{DBLP:conf/icassp/HuntB96, DBLP:journals/speech/ZenTB09, DBLP:journals/jmlr/PratapTSTBKENVF24} which often flattens prosody, CosyVoice enables precise control over accent and intonation via natural language prompts.

We implement a two-stage Rewrite-then-Synthesize strategy to bridge the `long-tail dialect gap': (1) \textit{Lexical Adaptation:} We employ a LLM (Qwen3-Instruct \cite{DBLP:journals/corr/abs-2505-09388}) to structurally rewrite standard text into dialectal vernacular (e.g., converting Mandarin vocabulary to distinct Cantonese or Northeastern lexical forms) while preserving semantic meaning. (2) \textit{Instruction-Aware Synthesis:} The rewritten text is processed by CosyVoice3 with dialect-specific instruction prompts, which injects specific dialectal characteristics—such as the tonal variations of Cantonese or the retroflexion of Northern Mandarin—into the text.
We generate fine-grained dialectal speech for 19 distinct Chinese regional variants, covering major linguistic groups (e.g., Cantonese, Wu, Minnan) and Mandarin accents (e.g., Sichuan, Dongbei, Tianjin).
This approach yields highly expressive audio that preserves the semantic content of the original text while introducing realistic phonological shifts characteristic of regional speakers. 
Our experiments (analyzed in Section~\ref{sec:experiments}) demonstrate that this generates data with sufficient discriminative power to serve as a valid proxy for human speech, bridging the data scarcity gap.
Please refer to \href{https://github.com/YoungSeng/PolySpeech-100}{our project repository} for audio samples.

  
  

\subsubsection{Track 3: Neural Synthesis for Long-Tail Languages} 

For low-resource languages where human data is inaccessible, we utilize high-fidelity Neural TTS. 
This approach is grounded in two key premises: First, TTS augmentation is essential for covering the linguistic long tail, a standard practice in recent multilingual evaluations~\cite{DBLP:conf/asru/KimKGGK21, DBLP:journals/corr/abs-2509-15373, DBLP:conf/acl/BarteldsSMJ023}. Second, subjective evaluations indicate that modern neural synthesis \footnote{\url{https://github.com/resemble-ai/chatterbox}} achieves MOS (Mean Opinion Score) parity with human recordings in clean settings~\cite{DBLP:journals/corr/abs-2406-02430, DBLP:journals/corr/abs-2509-08753, DBLP:journals/corr/abs-2506-21619}. 
Furthermore, this synthetic approach offers superior controllability; it allows us to systematically inject environmental noise and reverberation to assess model robustness, a factor we analyze extensively in the experimental section.
To extend coverage to under-represented languages (e.g., Zulu, Maltese, Lao, and various Arabic dialects), we employ high-fidelity neural text-to-speech (TTS) via the Edge-TTS engine \footnote{\url{https://github.com/rany2/edge-tts}}. We developed a rigorous mapping protocol to align FLORES-200 language codes with available neural voice locales. To mitigate speaker overfitting and improve model robustness, we enforce speaker diversity by randomizing voice profiles (Male/Female) across different samples \cite{shi2024predictive}.    

\subsection{Quality Assurance Protocol}

To ensure the reliability of PolySpeech-100, we implemented a rigorous three-step validation protocol: 
(1) \textit{Physical Integrity Verification}: We perform automated scanning of the generated corpus to detect file corruption. This includes checking for valid header structures, appropriate bit rates, and duration thresholds to filter out truncated or silent files. The dataset structure is normalized to ensure every sample contains a complete Q\&A chain (Passage, Question, and four Options).
(2) \textit{Semantic Consistency via ASR Loop \cite{DBLP:journals/corr/abs-2509-18004, DBLP:journals/corr/abs-2509-03959}}:
To guarantee that the synthetic speech remains intelligible and faithful to the source text, we introduce an Automatic Speech Recognition (ASR) verification loop.
To be specific, we utilized several tools (e.g., Qwen3-ASR \cite{shi2026qwen3}, SenseVoice \cite{DBLP:journals/corr/abs-2407-04051}, Whisper \cite{DBLP:conf/icml/RadfordKXBMS23}, TeleASR \cite{chen2024telespeechpt}) to transcribe synthetic audio back to text. We compute the Word Error Rate (WER) and Character Error Rate (CER) between the source text and the ASR transcription. Samples exceeding a strict error threshold are flagged as semantically distinct (indicating synthesis failure or severe hallucination) and are automatically regenerated or excluded.
This ensures that PolySpeech-100 measures speech understanding capabilities rather than robustness to poor audio generation.
(3) \textit{Human-in-the-Loop Validation}: To complement automated metrics, we conducted a manual spot-check on the dialectal samples \cite{ruan2025cpws}. Native speakers verified the authenticity of prosody and lexical usage, ensuring the dataset meets the quality standards required for academic benchmarking. 
Besides, we compared model performance on our data against real human recordings, observing a strong correlation ($r=0.83$), details are provided in Appendix.

\section{Benchmark Experiments}
\label{sec:experiments}

In this section, we evaluate 22 state-of-the-art audio systems on PolySpeech-100. We analyze their performance across high-resource languages, generated dialects, and long-tail languages.

\begin{table}[t]
\centering
\caption{Summary of Performance on PolySpeech-100. We categorize the languages into 3 groups: High-Res (10 major languages, e.g., EN, ZH), CN-Dialect (19 Chinese regional
dialects), and Low-Res (81 long-tail languages). Detailed definitions and full results are provided in Appendix C. \textbf{Bold} denotes the best result, and \underline{underline} denotes the second-best result. Note the significant gap between the SOTA closed-source model and others in Low-Res scenarios.}
\label{tab:key_results}

\small 
\resizebox{\linewidth}{!}{%
\begin{tabular}{llcccc}
\toprule

\multicolumn{2}{l}{\textbf{Model}} & \textbf{Overall} & \textbf{High-Res} & \textbf{CN-Dialect} & \textbf{Low-Res} \\ 
\midrule

\rowcolor{gray!12} 
\multicolumn{6}{l}{\textit{Closed-Source Models}} \\ 
\multicolumn{2}{l}{Gemini-3-flash \cite{google2025gemini3flash}} & \textbf{85.30} & \textbf{94.26} & \textbf{83.54} & \textbf{84.61} \\
\multicolumn{2}{l}{GPT-Audio-mini \cite{openai2025gptaudiomini}} & \underline{56.63} & 83.56 & 55.58 & \underline{53.56} \\ 

\addlinespace[1.0ex] 
\rowcolor{gray!12} 
\multicolumn{6}{l}{\textit{Open-Source E2E Models: {Speech+Text $\rightarrow$ Text}}} \\
\multicolumn{2}{l}{Fun-Audio-Chat \cite{DBLP:journals/corr/abs-2512-20156}} & 52.88 & 84.82 & 77.06 & 43.26 \\
\multicolumn{2}{l}{Qwen2.5-Omni \cite{DBLP:journals/corr/abs-2503-20215}} & 50.89 & \underline{84.94} & \underline{78.61} & 40.18 \\
\multicolumn{2}{l}{MiniCPM-o 4.5 \cite{DBLP:journals/corr/abs-2604-27393}} & 50.57 & 78.25 & 59.24 & 45.12 \\
\multicolumn{2}{l}{Audio-Omni \cite{DBLP:journals/corr/abs-2604-10708}} & 46.62 & {77.50} & {71.95} & 36.86 \\
\multicolumn{2}{l}{MiMo-Audio \cite{DBLP:journals/corr/abs-2512-23808}} & 43.51 & 61.09 & 76.03 & 33.72 \\
\multicolumn{2}{l}{Step-Audio-2 \cite{DBLP:journals/corr/abs-2507-16632}} & 40.52 & 60.44 & 68.17 & 31.57 \\
\multicolumn{2}{l}{Qwen2-Audio \cite{DBLP:journals/corr/abs-2407-10759}} & 25.94 & 26.59 & 25.40 & 25.99 \\

\addlinespace[0.5ex] 
\rowcolor{gray!12} 
\multicolumn{6}{l}{\textit{Open-Source E2E Models: {Speech $\rightarrow$ Text} }} \\       
\multicolumn{2}{l}{Covo-Audio \cite{DBLP:journals/corr/abs-2602-09823}} & 24.78 & 26.68 & 18.75 & 25.96 \\
\multicolumn{2}{l}{PersonaPlex \cite{DBLP:journals/corr/abs-2602-06053}} & 23.67 & 24.25 & 22.43 & 23.89 \\
\multicolumn{2}{l}{Mini-Omni \cite{DBLP:journals/corr/abs-2408-16725}} & 21.83 & 17.64 & 22.91 & 22.10 \\
\multicolumn{2}{l}{LLaMA-Omni2 \cite{DBLP:conf/iclr/FangGZMZ025}} & 21.88 & 22.87 & 20.26 & 22.14 \\
\multicolumn{2}{l}{Moshi \cite{DBLP:journals/corr/abs-2410-00037}} & 20.96 & 25.46 & 17.78 & 21.15 \\ 

\addlinespace[1.0ex] 
\rowcolor{gray!12} 
\multicolumn{6}{l}{\textit{Cascaded Pipelines (ASR+LLM)}} \\

\multirow{4}{*}{\shortstack[l]{Whisper-v3 \\ \cite{DBLP:conf/icml/RadfordKXBMS23}}} 
 & + Qwen2.5 \cite{DBLP:journals/corr/abs-2412-15115} & 53.86 & 83.74 & 62.62 & 48.12 \\
 & + Qwen3 \cite{DBLP:journals/corr/abs-2505-09388} & 51.56 & 80.17 & 59.82 & 46.09 \\
 & + Llama-3.1 \cite{DBLP:journals/corr/abs-2407-21783} & 43.59 & 62.74 & 46.86 & 40.46 \\
 & + Llama-3.2 \cite{DBLP:journals/corr/abs-2407-21783} & 39.01 & 59.20 & 44.98 & 35.12 \\

\addlinespace[1.0ex]

\multirow{4}{*}{\shortstack[l]{Qwen3-ASR \\ \cite{shi2026qwen3}}} 
 & + Qwen3 \cite{DBLP:journals/corr/abs-2505-09388} & 52.66 & 84.28 & 73.93 & 43.77 \\
 & + Qwen2.5 \cite{DBLP:journals/corr/abs-2412-15115} & 52.29 & 81.80 & 72.36 & 43.94 \\
 & + Llama-3.1 \cite{DBLP:journals/corr/abs-2407-21783} & 45.00 & 66.81 & 58.31 & 39.18 \\ 
 & + Llama-3.2 \cite{DBLP:journals/corr/abs-2407-21783} & 38.21 & 57.04 & 49.36 & 33.27\\ 
\bottomrule
\end{tabular}%
}
\end{table}

\begin{figure*}[h]
  \centering
  \includegraphics[width=\linewidth]{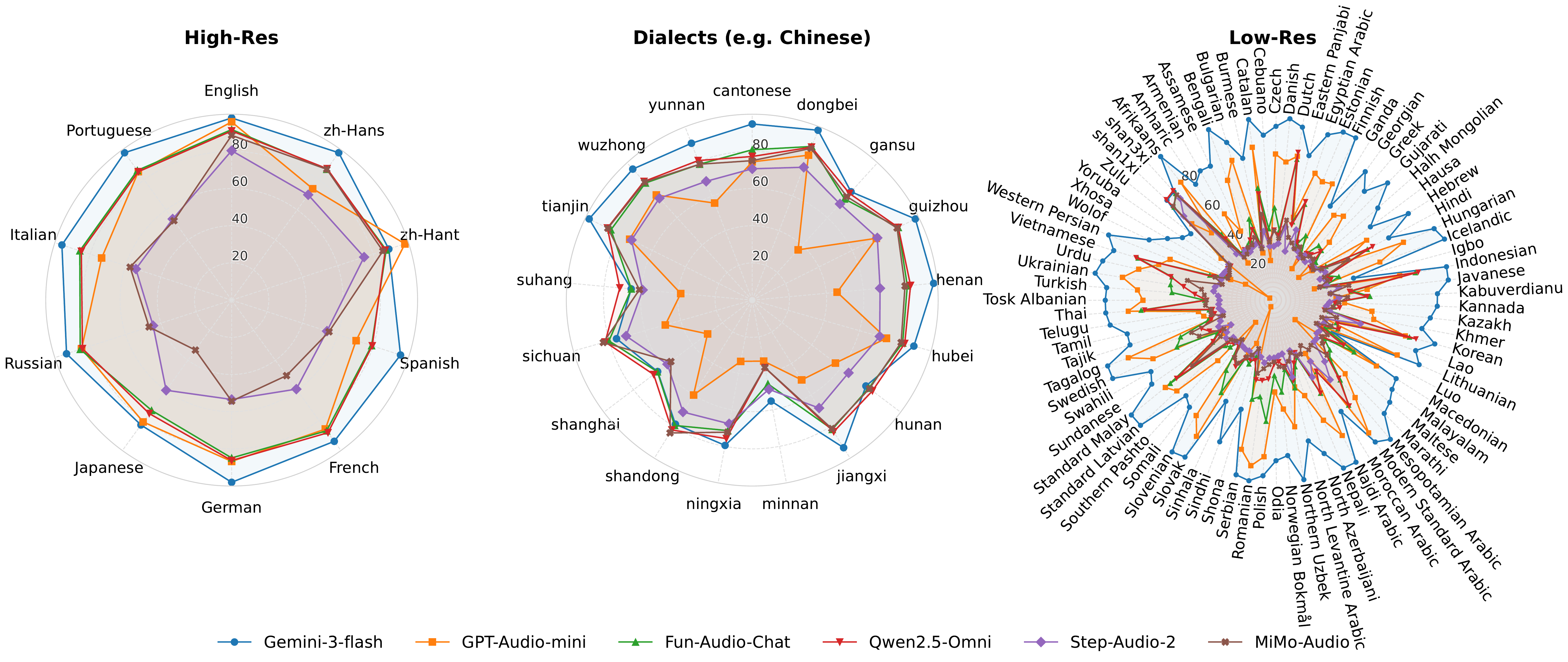}
  \caption{Performance benchmarking of various audio models across diverse linguistic scenarios. The radar charts illustrate comparisons in High-Resource languages (left), Chinese Dialects (middle), and Low-Resource languages (right). The visualization highlights the capabilities of models such as Gemini-3-flash and Fun-Audio-Chat, demonstrating performance variances particularly in dialectal and low-resource contexts.}
  \label{fig:5}
\end{figure*}

\subsection{Experimental Setup}
We categorize the evaluated models into two primary groups. The first group consists of End-to-End (E2E) models that process audio inputs directly without requiring intermediate text transcription. Within this category, we evaluated closed-source models such as \texttt{Gemini-3.0-Flash-Preview} \cite{google2025gemini3flash} and \texttt{gpt-audio-mini-2025-\allowbreak12-15} \cite{openai2025gptaudiomini}. We also selected eight open-source models, including \texttt{Qwen2.5-Omni-7B} \cite{DBLP:journals/corr/abs-2503-20215}, \texttt{Fun-Audio-Chat-8B} \cite{DBLP:journals/corr/abs-2512-20156}, \texttt{Step-Audio-2-\allowbreak Mini} \cite{DBLP:journals/corr/abs-2507-16632}, \texttt{MiMo-Audio-7B} \cite{DBLP:journals/corr/abs-2512-23808}, \texttt{Qwen2-Audio-7B-Instruct} \cite{DBLP:journals/corr/abs-2407-10759}, \texttt{LLaMA-Omni2-7B-Bilingual} \cite{DBLP:journals/corr/abs-2505-02625}, \texttt{Mini-Omni} \cite{DBLP:journals/corr/abs-2408-16725}, and \texttt{Moshi} \cite{DBLP:journals/corr/abs-2410-00037}.

The second group comprises Cascade Systems, serving as baselines that combine ASR modules with LLM. For the ASR frontend, we utilized \texttt{Whisper-Large-v3} \cite{DBLP:conf/icml/RadfordKXBMS23} and \texttt{Qwen3-ASR-1.7B} \cite{shi2026qwen3}. These are paired with LLM backends as \texttt{Qwen3-4B-Instruct-\allowbreak2507} \cite{DBLP:journals/corr/abs-2505-09388}, \texttt{Qwen2.5-7B-Instruct} \cite{DBLP:journals/corr/abs-2412-15115}, \texttt{Llama-3.2-3B-Instruct} and \texttt{Llama-\allowbreak3.1-8B-Instruct} \cite{DBLP:journals/corr/abs-2407-21783}.

Regarding the prompting strategy, we adapted our approach based on the specific capabilities of each system. For models that support system instructions (e.g., Qwen2.5-Omni, Gemini), we utilized a text system prompt to define the task. However, for models that do not support system instructions—specifically (e.g., \texttt{LLaMA-Omni2}, \texttt{Moshi}, \texttt{Mini-Omni}), we converted the instruction into speech using TTS and concatenated it to the beginning of the input audio. The exact prompts used are provided in the Appendix.

\subsection{Main Results}
\label{sec:main_results}

\subsubsection{Overall Performance Analysis}

As shown in Table \ref{tab:key_results}, in the general evaluation across all 100+ languages, Gemini-3-flash achieves state-of-the-art performance, achieving an overall accuracy of 85.30\% across all languages. It maintains consistent performance across both high-resource and low-resource settings, demonstrating remarkable robustness. 
In contrast, OpenAI's GPT-Audio-mini lags significantly with an overall score of 56.63\%, placing it in a similar tier to the best open-source models rather than leading the field.

Among open-source E2E models, Fun-Audio-Chat (52.88\%) and Qwen2.5-Omni (50.89\%) emerge as the top performers. Their performance is highly competitive with, and in some aspects surpasses, the strong baseline of cascaded pipelines (e.g., Whisper-v3 + Qwen2.5 at 53.86\%). However, models that rely on `Prompt-to-Speech' concatenation due to a lack of system prompt support (specifically Mini-Omni, LLaMA-Omni2, and Moshi) collapsed catastrophically. Their scores hovered around 20-22\%, which is below the random guess threshold (25\%) for four-choice questions. This finding strongly suggests that current audio-language models struggle to distinguish instruction from content when they are mixed in the same audio stream without textual guidance.

\subsubsection{Performance on Dialects} 

The evaluation on 19 Chinese regional dialects (and 6 Arabic variants, see Appendix \ref{sec:arab_case_study} for a detailed case study) reveals striking insights. While Gemini-3-flash remains the top scorer (83.54\%), open-source E2E models exhibit a remarkable ``native understanding'' advantage over traditional pipelines and even GPT-Audio-mini.
Specifically, Qwen2.5-Omni and Fun-Audio-Chat achieved accuracy scores of 78.61\% and 77.06\%, respectively. This performance significantly outperforms GPT-Audio-mini (55.58\%) by a margin of over 20 percentage points. Furthermore, these E2E models surpass the classic Whisper-v3 + Qwen2.5 pipeline (62.62\%). This indicates that general-purpose ASR systems like Whisper often lose critical semantic information when transcribing heavily accented dialects into text. Although using a specialized ASR (Qwen3-ASR) improves pipeline performance to 73.93\%, the best E2E model (Qwen2.5-Omni) still holds the lead. This empirically validates the hypothesis that E2E architectures can leverage paralinguistic cues and latent representations to bridge the gap in dialectal understanding better than text-mediated systems.
For a detailed case analysis, please refer to Appendix~\ref{sec:appendix_case}.

Besides, we observe a clear hierarchy in model performance based on linguistic distance. As shown in Fig. \ref{fig:5}, on Standard Mandarin (\texttt{zho\_Hans}) and Northern dialects such as \texttt{Tianjin} and \texttt{Dong\-bei}, models like Qwen2.5-Omni achieve greater than 85\% accuracy. Dialects with distinct tones but shared vocabulary, like \texttt{Sichuan}, accuracy remains robust at approximately 80\%. However, performance degrades significantly on distinct dialect families like \texttt{Shanghai} (Wu) and \texttt{Cantonese} (Yue). For example, Qwen2.5-Omni drops from 87.4\% (Mandarin) to 66.4\% (Shanghai), validating our synthesized data in challenging the models. The gap in Arabic varieties is even more severe. While models perform well on Modern Standard Arabic (MSA), they struggle with regional variants; Fun-Audio-Chat scores 69.7\% on MSA but drops to 45.1\% on Moroccan Arabic. This highlights a standard language bias in current training datasets, where dialects are underrepresented compared to formal speech.

\subsubsection{Robustness on Low-Resource Languages} 
The `Low-Res' category, comprising 81 long-tail languages, proved to be the most challenging testbed. We observed a universal performance degradation, but the severity varied drastically across models. Gemini-3-flash exhibited exceptional robustness, maintaining a high accuracy of 84.61\%, which is nearly identical to its high-resource performance.

In contrast, all other models experienced a sharp decline. GPT-Audio-mini dropped to 53.56\%, while the best open-source E2E models fell into the 30-43\% range (e.g., Fun-Audio-Chat at 43.26\%). Interestingly, in this specific domain, the Whisper-v3 + Qwen2.5 pipeline (48.12\%) slightly outperformed the best open-source E2E models. This suggests that for ultra-low-resource languages, the massive multilingual pre-training of Whisper's encoder currently provides a more stable foundation than the audio encoders used in current open-source E2E models. Nevertheless, the substantial gap between Gemini and all other models highlights that low-resource speech understanding remains the primary frontier for the open-source community to conquer.

\paragraph{Efficiency Analysis}
Beyond accuracy, we analyzed the computational cost by standardizing the total inference time for the Poly\-Speech-100 benchmark (88,000 samples). Fun-Audio-Chat is the most efficient model (11 hours). This high speed makes it ideal for real-time applications where latency is critical. LLaMA-Omni2 (15.8 hours), MiMo-Audio (18.2 hours), and Qwen2.5-Omni (21.3 hours) show moderate efficiency. 
These architectures, while powerful, result in higher costs due to their streaming mechanisms.
Step-Audio-2 takes slightly longer at 30.0 hours. This suggests potential optimization bottlenecks in its inference implementation or a heavier encoder architecture. 
In contrast, Mini-Omni, Qwen2-Audio and Moshi are significantly slower (109.5, 141.0, and 200+ hours respectively). Although these three models possess chat capabilities, the long duration of each input audio segment likely causes the complexity to increase exponentially, causing high latency.

\begin{figure}[!htbp]
    \centering
    
    \begin{subfigure}[b]{1.0\linewidth}
        \centering
        \includegraphics[width=\linewidth]{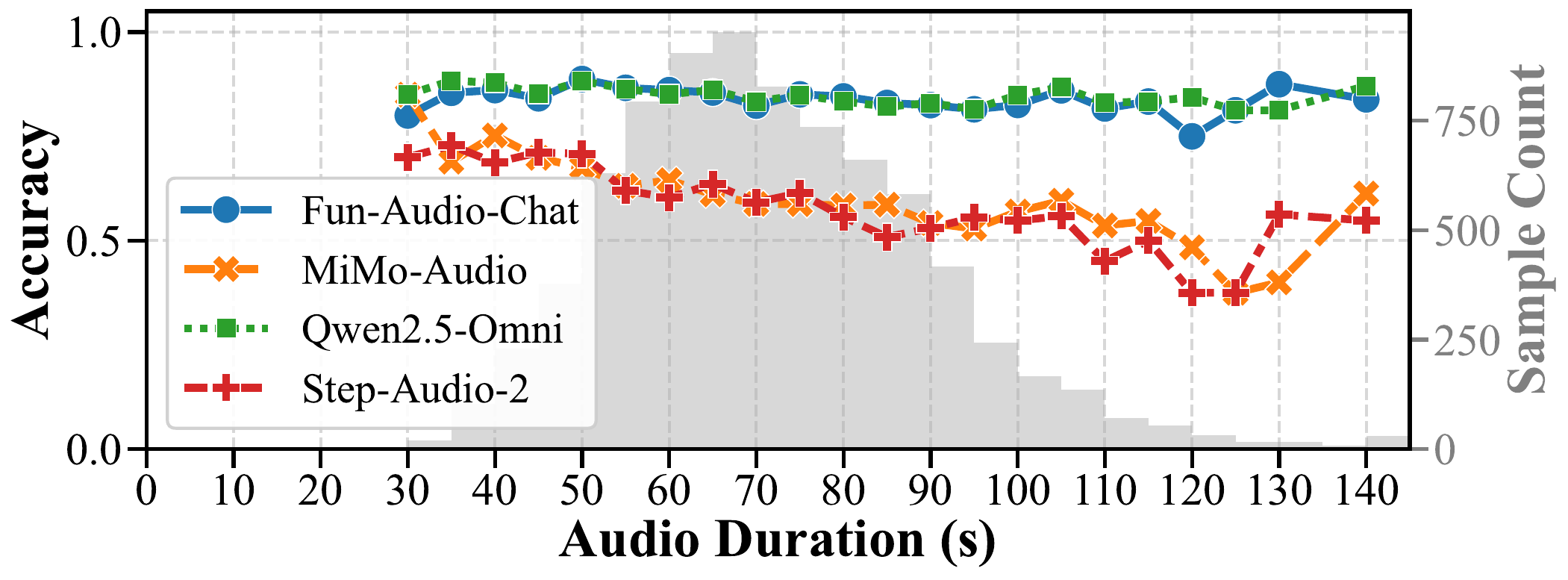} 
        \caption{High-Resource Languages}
        \label{fig:sub1}
    \end{subfigure}
    
    \vspace{1em} 
    
    \begin{subfigure}[b]{1.0\linewidth}
        \centering
        \includegraphics[width=\linewidth]{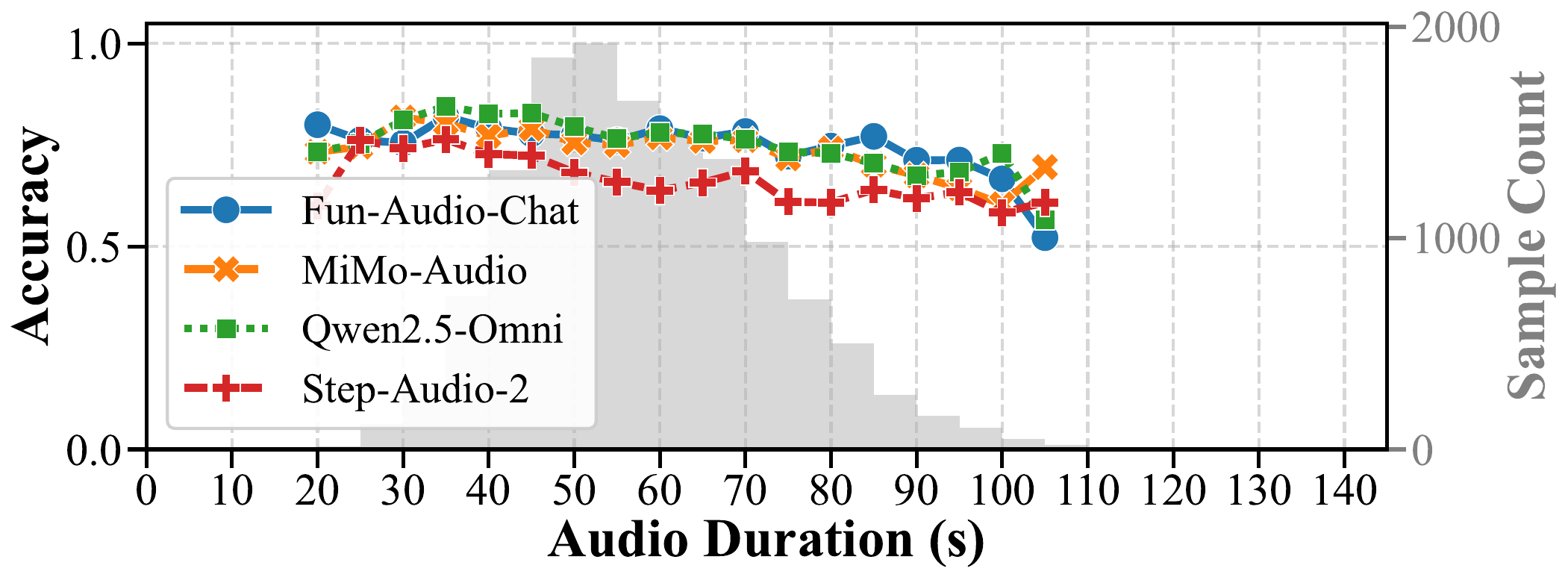}
        \caption{Chinese Dialects}
        \label{fig:sub2}
    \end{subfigure}
    
    \vspace{1em} 
    
    \begin{subfigure}[b]{1.0\linewidth}
        \centering
        \includegraphics[width=\linewidth]{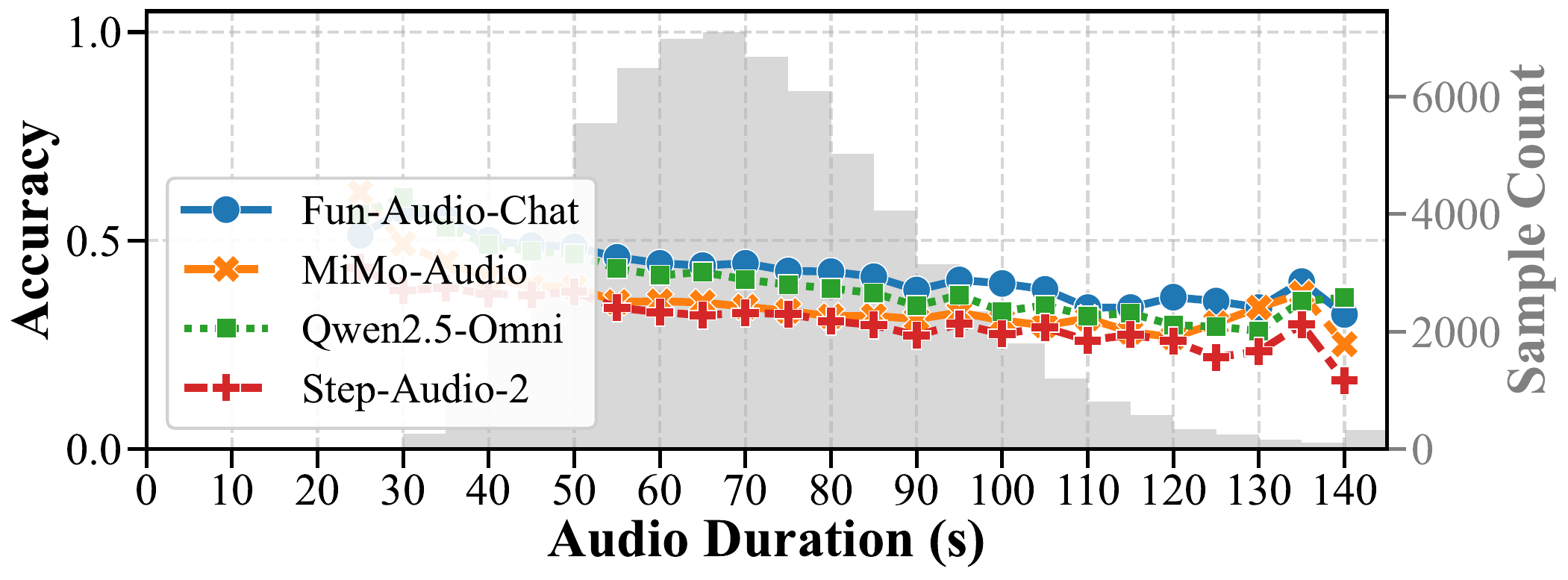}
        \caption{Long-Tail Languages}
        \label{fig:sub3}
    \end{subfigure}
    
    \caption{Performance analysis of 4 Speech-LLMs across varying audio durations. The colored line charts (left axis) illustrate the accuracy of each model, while the background gray histograms (right axis) indicate the number of test samples per duration interval (minimum samples threshold$=$15, cutoff=140s). This highlights the correlation between data sparsity (indicated by lower histogram bars) and model performance degradation in long-form audio segments.}
    \label{fig:6}
    
\end{figure}

\begin{table*}[t]
\centering
\caption{\textbf{Comprehensive Performance Analysis with Heatmap.} We report accuracy (\%) and the absolute performance drop/gain (in parentheses) relative to the Base condition. The cell background color indicates the magnitude and direction of the change: \textbf{\textcolor{captionRed}{Red}} shades denote performance degradation, while \textbf{\textcolor{captionGreen}{Green}} shades denote improvement. Darker shades indicate larger changes.}
\label{tab:merged_performance_heatmap}
\resizebox{\textwidth}{!}{%
\begin{tabular}{l|l|c|cc|cc|cc}
\toprule
\multirow{2}{*}{\textbf{Group}} & \multirow{2}{*}{\textbf{Model}} & \textbf{Base} & \multicolumn{2}{c|}{\textbf{Noise Robustness}} & \multicolumn{2}{c|}{\textbf{Speed Robustness}} & \multicolumn{2}{c}{\textbf{Strategy Impact}} \\
 & & \textbf{(Zero-Shot)} & \textbf{Noise-Low} & \textbf{Noise-High} & \textbf{Speed-Slow} & \textbf{Speed-Fast} & \textbf{CoT Prompting} & \textbf{3-Shot Context} \\ \midrule
\multirow{4}{*}{\textbf{High-Res}} 
& Fun-Audio-Chat \cite{DBLP:journals/corr/abs-2512-20156} & 84.82 & \ccr{81.10 (-3.7)} & \ccRR{65.73 (-19.1)} & \ccr{81.35 (-3.5)} & \ccr{82.24 (-2.6)} & \ccR{75.36 (-9.46)} & \ccr{82.69 (-2.13)} \\
& MiMo-Audio \cite{DBLP:journals/corr/abs-2512-23808} & 61.09 & \ccr{58.46 (-2.6)} & \ccRR{49.29 (-11.8)} & \ccN{61.10 (+0.0)} & \ccr{57.88 (-3.2)} & \ccRR{50.11 (-10.98)} & \ccR{54.04 (-7.05)} \\
& Qwen2.5-Omni \cite{DBLP:journals/corr/abs-2503-20215} & 84.94 & \ccN{\textbf{83.96 (-1.0)}} & \ccR{\textbf{75.90 (-9.0)}} & \ccr{\textbf{83.67 (-1.3)}} & \ccN{\textbf{85.15 (+0.2)}} & \ccRR{74.06 (-10.88)} & \ccg{86.64 (+1.70)} \\
& Step-Audio-2 \cite{DBLP:journals/corr/abs-2507-16632} & 60.44 & \ccN{61.09 (+0.7)} & \ccRR{47.37 (-13.1)} & \ccN{59.49 (-1.0)} & \ccN{59.61 (-0.8)} & \ccG{67.54 (+7.10)} & \ccRR{45.39 (-15.05)} \\ \midrule
\multirow{4}{*}{\textbf{CN-Dialect}} 
& Fun-Audio-Chat \cite{DBLP:journals/corr/abs-2512-20156} & 77.06 & \ccr{73.88 (-3.2)} & \ccRR{66.37 (-10.7)} & \ccr{73.65 (-3.4)} & \ccR{68.27 (-8.8)} & \ccN{77.14 (+0.08)} & \ccN{76.55 (-0.51)} \\
& MiMo-Audio \cite{DBLP:journals/corr/abs-2512-23808} & 76.03 & \ccN{75.96 (-0.1)} & \ccR{68.03 (-8.0)} & \ccN{75.73 (-0.3)} & \ccr{74.12 (-1.9)} & \ccN{76.32 (+0.29)} & \ccN{75.92 (-0.11)} \\
& Qwen2.5-Omni \cite{DBLP:journals/corr/abs-2503-20215} & 78.61 & \ccN{\textbf{77.76 (-0.9)}} & \ccR{\textbf{70.50 (-8.1)}} & \ccr{\textbf{75.89 (-2.7)}} & \ccr{\textbf{77.17 (-1.4)}} & \ccRR{68.32 (-10.29)} & \ccN{77.82 (-0.79)} \\
& Step-Audio-2 \cite{DBLP:journals/corr/abs-2507-16632} & 68.17 & \ccr{64.13 (-4.0)} & \ccRR{55.62 (-12.6)} & \ccr{66.65 (-1.5)} & \ccr{66.05 (-2.1)} & \ccr{66.72 (-1.45)} & \ccR{59.53 (-8.64)} \\ \midrule
\multirow{4}{*}{\textbf{Low-Res}} 
& Fun-Audio-Chat \cite{DBLP:journals/corr/abs-2512-20156} & 43.26 & \ccR{37.51 (-5.8)} & \ccRR{30.95 (-12.3)} & \ccR{38.24 (-5.0)} & \ccR{38.09 (-5.2)} & \ccr{39.06 (-4.20)} & \ccr{41.52 (-1.74)} \\
& MiMo-Audio \cite{DBLP:journals/corr/abs-2512-23808} & 33.72 & \ccr{30.92 (-2.8)} & \ccR{28.10 (-5.6)} & \ccr{31.40 (-2.3)} & \ccr{30.61 (-3.1)} & \ccr{31.75 (-1.97)} & \ccr{30.29 (-3.43)} \\
& Qwen2.5-Omni \cite{DBLP:journals/corr/abs-2503-20215} & 40.18 & \ccr{37.08 (-3.1)} & \ccR{32.97 (-7.2)} & \ccR{34.69 (-5.5)} & \ccr{36.53 (-3.7)} & \ccR{33.37 (-6.81)} & \ccN{40.57 (+0.39)} \\
& Step-Audio-2 \cite{DBLP:journals/corr/abs-2507-16632} & 31.57 & \ccr{29.51 (-2.1)} & \ccr{28.72 (-2.9)} & \ccr{28.99 (-2.6)} & \ccr{29.27 (-2.3)} & \ccN{31.54 (-0.03)} & \ccR{22.88 (-8.69)} \\ \bottomrule
\end{tabular}%
}
\end{table*}

\subsection{Robustness Analysis}
\label{sec:robustness}

Real-world speech interaction rarely occurs in studio-quality conditions. To evaluate the resilience of Speech-LLMs against acoustic distortions and their operational efficiency, we conducted a stress test on PolySpeech-100. We selected four representative open-source models—\textit{Fun-Audio-Chat}, \textit{MiMo-Audio}, \textit{Qwen2.5-Omni}, and \textit{Step-Audio-2}—and evaluated them under five acoustic conditions across three language groups.
We implemented an automated augmentation pipeline. The original recordings serve as the clean baseline. To simulate real-world interference, we add Gaussian white noise at two distinct levels: a low setting with a signal-to-noise ratio (SNR) of approximately 20dB to mimic a quiet office, and a high setting with roughly 0-5dB SNR to represent a noisy street environment (using peak normalization). Additionally, we adjust the audio speed to create temporal variations. This includes a slow version at 0.8x the original speed and a fast version at 1.2x, which modifies both the tempo and pitch of the input.
Besides, we recorded the total wall-clock time required to complete the full benchmark (all 100+ languages) on a single NVIDIA 4090 GPU. This provides a direct measure of inference throughput.

Table \ref{tab:merged_performance_heatmap} presents the performance stability across different language groups. Values in parentheses indicate the absolute accuracy drop compared to the \textit{Base} (Clean) condition.

\subsubsection{Sensitivity to Environmental Noise}
Table \ref{tab:merged_performance_heatmap} shows, High-Res languages suffer significant degradation under high noise. For instance, \textit{Fun-Audio-Chat} drops by 19.1\% in the High-Res group. In contrast, Qwen2.5-Omni demonstrates superior resilience, dropping only 9.0\%. This suggests that the streaming-oriented pre-training of Qwen2.5-Omni likely included large-scale noisy data, making it more robust to acoustic interference.
Interestingly, the performance drop on Dialects (e.g., Qwen2.5-Omni drop of 8.1\%) is comparable to High-Res languages. This indicates that the dialectal features learned by these models are robust and not merely fragile overfitting to clean TTS data.
In the Low-Res group, absolute performance is already low ($\sim$30-40\%). Consequently, the impact of noise appears numerically smaller (e.g., Step-Audio-2 drops only 2.9\%), but this is largely due to the ``floor effect''—performance cannot drop much further below random guessing (25\%). However, \textit{Fun-Audio-Chat} still loses over 12\% accuracy, highlighting that its acoustic encoder is less stable on unfamiliar phonemes when noise is present.

\begin{figure*}[t!]
  \centering
  \includegraphics[width=\linewidth]{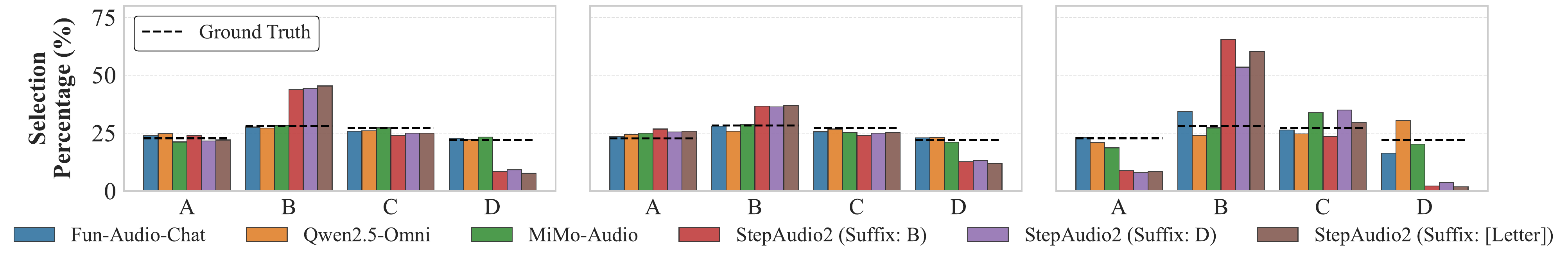}
  \caption{Analysis of prediction bias and prompt sensitivity across three linguistic groups. The bar charts display the selection rate (\%) of each option (A, B, C, D) for six model variations: Fun-Audio, Qwen2.5-Omni, MiMo-Audio, and StepAudio2 with three different system prompt suffixes. The black dashed lines represent the ground truth distribution. Significant deviations from the dashed lines indicate a model's preference bias towards specific options.}
  \label{fig:7}
\end{figure*}

\subsubsection{Sensitivity to Speaking Rate}
Models generally handled speed variations better than noise.
However, for \textit{Fun-Audio-Chat}, the fast speed setting caused a notable drop in dialect performance (-8.8\%), suggesting that temporal compression makes accent recognition significantly harder for some architectures.

\subsubsection{Analysis of Audio Duration Robustness}

To evaluate the robustness of Speech-LLMs across different input lengths, we analyze model performance with respect to audio duration as shown in Figure~\ref{fig:6}. While a general negative correlation exists between audio duration and model accuracy, high-resource language scenarios remain largely stable; specifically, Fun-Audio-Chat and Qwen2.5-Omni exhibit remarkable stability, maintaining high accuracy (near $0.8-0.9$) even as the duration extends to 140 seconds, whereas MiMo-Audio and Step-Audio-2 show significant drops for longer segments. Conversely, in long-tail languages, although overall accuracy is significantly lower, Fun-Audio-Chat consistently achieves leading performance across almost all intervals. It is worth noting that Chinese dialects generally exhibit shorter durations compared to other languages, and the apparent accuracy fluctuations in the 120–140s interval are primarily artifacts of limited sample sizes. 

\subsubsection{Analysis of Prediction Bias}

To investigate model reliability, we analyze the prediction bias and prompt sensitivity as illustrated in Figure \ref{fig:7}. We observe distinct behaviors among the models: while \textit{Fun-Audio-Chat}, \textit{Qwen2.5-Omni}, and \textit{MiMo-Audio} demonstrate robust performance with selection distributions aligning closely with the ground truth, the \textit{Step-Audio-2} variants exhibit a severe preference bias. Specifically, \textit{Step-Audio-2} disproportionately selects option B, a trend that remains consistent regardless of the system prompt suffix applied (refer to the Appendix for specific prompts and modifications). This persistence indicates an inherent model bias rather than simple prompt sensitivity, contrasting sharply with the balanced distribution observed in the other models.

\subsection{Advanced Reasoning Capabilities}
\label{sec:advanced_reasoning}

To investigate the advanced reasoning potential of Speech-LLMs beyond standard zero-shot accuracy, we employed two paradigms: Chain-of-Thought (CoT) prompting, where system prompts were modified to explicitly request reasoning steps prior to the final answer, and Few-Shot (In-Context) Learning, which involved prepending a 3-turn audio segment containing [Passage, Question, Answer] examples to the input; the impact of these techniques across the three language groups is summarized in Table \ref{tab:merged_performance_heatmap}.

\subsubsection{Divergent Impact of Chain-of-Thought on Speech-LLMs}

Contrary to the consistent positive effect typically observed in text-based LLMs, our zero-shot experiments indicate that Chain-of-Thought prompting frequently exerted a negative impact on several current Speech-LLMs. Specifically, models such as
 \textit{Qwen2.5-Omni} and \textit{Fun-Audio-Chat} suffered a performance degradation of approximately 10\% in High-Resource languages. 
We hypothesize this stems from reasoning hallucinations, where the model is misled during the inference phase, and we refer readers to the Appendix \ref{sec:cot_failure_analysis} for a detailed qualitative analysis of these failure cases. A notable exception was \textit{Step-Audio-2}, which achieved a significant 7.1\% improvement in the same category, 
suggesting that its unique alignment training likely incorporated audio reasoning data enabling effective CoT reasoning.
To verify that this degradation is not merely an artifact of a specific prompt design, we conducted an ablation study using multiple alternative CoT templates (detailed in Appendix \ref{sec:cot_ablation}). The consistent performance drops observed across different prompt structures confirm that this is a fundamental modality alignment issue rather than prompt sensitivity.

\subsubsection{Ineffectiveness of Audio In-Context Learning}
The 3-Shot experiments highlighted limitations in current audio context processing, as most models failed to leverage the prepended examples. Specifically, \textit{Step-Audio-2} suffered a massive drop (e.g., -15.05\% in High-Resource languages), indicating that the long context likely exceeded its effective audio attention span and caused the model to `forget' the actual question; conversely, while \textit{Qwen2.5-Omni} showed slight positive transfer in High-Resource (+1.7\%) and Low-Resource (+0.39\%) settings—validating its stronger backbone—these gains were negligible compared to the computational cost of processing the extra audio tokens.

\section{Discussion}

The evaluation results from PolySpeech-100 reveal a fundamental limitation in current cascade systems (ASR + LLM), the information bottleneck created by converting speech into intermediate text. 
Our experiments demonstrate that text serves as a lossy compression of speech, where standard ASR normalization inadvertently strips away paralinguistic cues—such as tone, stress, and regional nuances—that are critical for semantic disambiguation in heavy dialects. 
In contrast, End-to-End models bypass this transcription phase by leveraging continuous latent representations, thereby retaining the acoustic fidelity necessary for complex reasoning. 
This evidence suggests that to achieve truly inclusive speech understanding, particularly for dialectal speakers and strictly oral languages (e.g., unwritten languages), the field must transition from a transcribe-then-read paradigm to native acoustic reasoning.

Unlike the consistent gains seen in text-based models, our study suggests that applying standard Chain-of-Thought (CoT) prompting to current Speech-LLMs can sometimes degrade performance and increase hallucinations, depending on the specific prompt design and model architecture.
This suggests a modality misalignment where intermediate text generation decouples the model from acoustic cues.
We attribute this failure to a fundamental modality alignment gap: current models are trained on direct transcription or immediate answering tasks, lacking speech-to-reasoning supervision to ground logic in acoustic features. Consequently, when forced to generate reasoning steps, the model decouples from the audio input and reverts to internal text priors, indicating that future work must shift focus toward training curricula that explicitly align audio representations with complex reasoning paths.

Despite the extensive scale of PolySpeech-100, we identify several meaningful directions for future enhancement. 
First, while our coverage is broad, there is an opportunity to extend support to under-represented languages—such as Tibetan (in Western China) and various indigenous languages in the Amazon and Pacific Islands—which may rely on raw recordings and self-supervised learning. Second, our current evaluation employs a multiple-choice format to prioritize precise speech comprehension; however, we recognize that authentic chat is inherently open-ended. Therefore, this work serves as a foundation for understanding, paving the way for future systems that support full-duplex interaction and complex turn-taking. Finally, our questions are designed with a neutral tone to isolate semantic accuracy, suggesting that subsequent iterations could incorporate paralinguistic signals \cite{DBLP:journals/corr/abs-2507-17563}, such as emotion and prosody, to move towards holistic social audio understanding.

\section{Conclusion}
In this paper, we introduced PolySpeech-100, a comprehensive benchmark designed to evaluate Speech-Large Language Models (Speech-LLMs) across 110 languages and dialects. By employing a hybrid data construction pipeline that combines human recordings with instruction-driven synthesis, we effectively addressed the scarcity of evaluation resources for regional dialects and low-resource languages. 
Our extensive experiments on 22 state-of-the-art models reveal that End-to-End architectures significantly outperform traditional ASR-Cascaded systems in understanding heavy dialects, demonstrating that direct audio processing captures rich acoustic information necessary for disambiguation, which is typically normalized out in textual transcription.
However, a substantial performance gap persists between closed-source and open-source models regarding low-resource languages. 
Additionally, we observed that standard text-based reasoning strategies, such as Chain-of-Thought, do not consistently yield improvements and can currently degrade audio performance for certain models, highlighting a need for better modality alignment.
We hope PolySpeech-100 establishes a rigorous standard to foster the development of next-generation, inclusive, and truly omni-capable speech systems.

\section{Acknowledgement}

This work was partially supported by the National Natural Science Foundation of China (No. 62293544, No. 62425117, No. 62506205), the Guangdong Basic and Applied Basic Research Foundation (No. 2026A1515011804), the China Postdoctoral Science Foundation (No. 2025T180426), the Postdoctoral Fellowship Program of CPSF (No. GZB20250393). We thank Xiaodong He and Youzheng Wu of JD.COM for their in-depth discussions during the research and development of PolySpeech-100.

\bibliographystyle{ACM-Reference-Format}
\bibliography{main_cleaned}

\appendix

\section{Data Construction and Validation}
\label{app:data_construction}


\subsection{Research Motivation and Dataset Design}
Recent Speech Large Language Models have achieved impressive understanding of spoken language, yet traditional evaluations focus primarily on high-resource languages and standard accents. This focus creates a performance gap regarding regional dialects that effectively bottlenecks real-world global deployment. 
To address the necessity for end-to-end processing across diverse linguistic variants, we introduce PolySpeech-100. This benchmark employs a data aggregation strategy combining high-quality human recordings with high-fidelity synthetic speech. 
We target 73 languages and 43 fine-grained dialects. Our extensive evaluation reveals that while performance on standard languages is strong, there is a significant weakness in processing regional dialects. This underscores the critical importance of evaluating diverse linguistic variants.

\subsection{Speech Synthesis and Model Landscape}
The construction of PolySpeech-100 relies on the observation that TTS has reached a level of fidelity comparable to human speech, like Orpheus-TTS \footnote{\url{https://github.com/canopyai/Orpheus-TTS}}, OmniVoice \cite{DBLP:journals/corr/abs-2604-00688}, MiMo-V2-TTS \cite{xiaomi2026mimov2tts}, and StepAudio 2.5 \cite{stepfun2024stepaudio}—which generate highly expressive and prosodically rich audio.
This high quality allows us to scale dialect coverage effectively. Beyond synthesis, it is crucial to acknowledge the evolving landscape of speech interaction models. 
While our benchmark focuses heavily on End-to-End Speech LLMs, traditional Cascade systems remain highly relevant. 
Recent work such as X-Talk and Voice Agents utilizing NVIDIA \footnote{\url{https://github.com/pipecat-ai/nemotron-january-2026}} demonstrates that cascading ASR, LLM, and TTS modules can achieve excellent latency and accuracy, particularly for streaming applications. Consequently, we posit that the future of voice interaction will not be dominated by a single architecture. Instead, the field will likely advance via a hybrid approach where End-to-End models and optimized Cascade systems develop in parallel, each addressing different trade-offs between semantic understanding and real-time response.

\subsection{Acoustic Diversity Analysis}
To assess task complexity, we analyzed the acoustic features of the entire PolySpeech-100 benchmark. We randomly sampled 100 utterances per language from the full dataset, extracted embeddings using the Wav2Vec2-XLS-R-300M model. 
The resulting t-SNE visualization reveals a highly entangled feature space without distinct boundaries between languages. This lack of clear clustering indicates that the acoustic diversity is high and that simple surface-level features are insufficient for discrimination. Consequently, a model cannot rely on basic pattern matching or acoustic heuristics to solve the tasks in PolySpeech-100. 
The absence of distinct clusters indicates that effectively disentangling these diverse linguistic signals requires more than simple acoustic heuristics; specifically, the overlapping distribution confirms the necessity for robust cross-lingual representations and deep semantic understanding.

\begin{figure}[h]
    \centering
    \includegraphics[width=\linewidth]{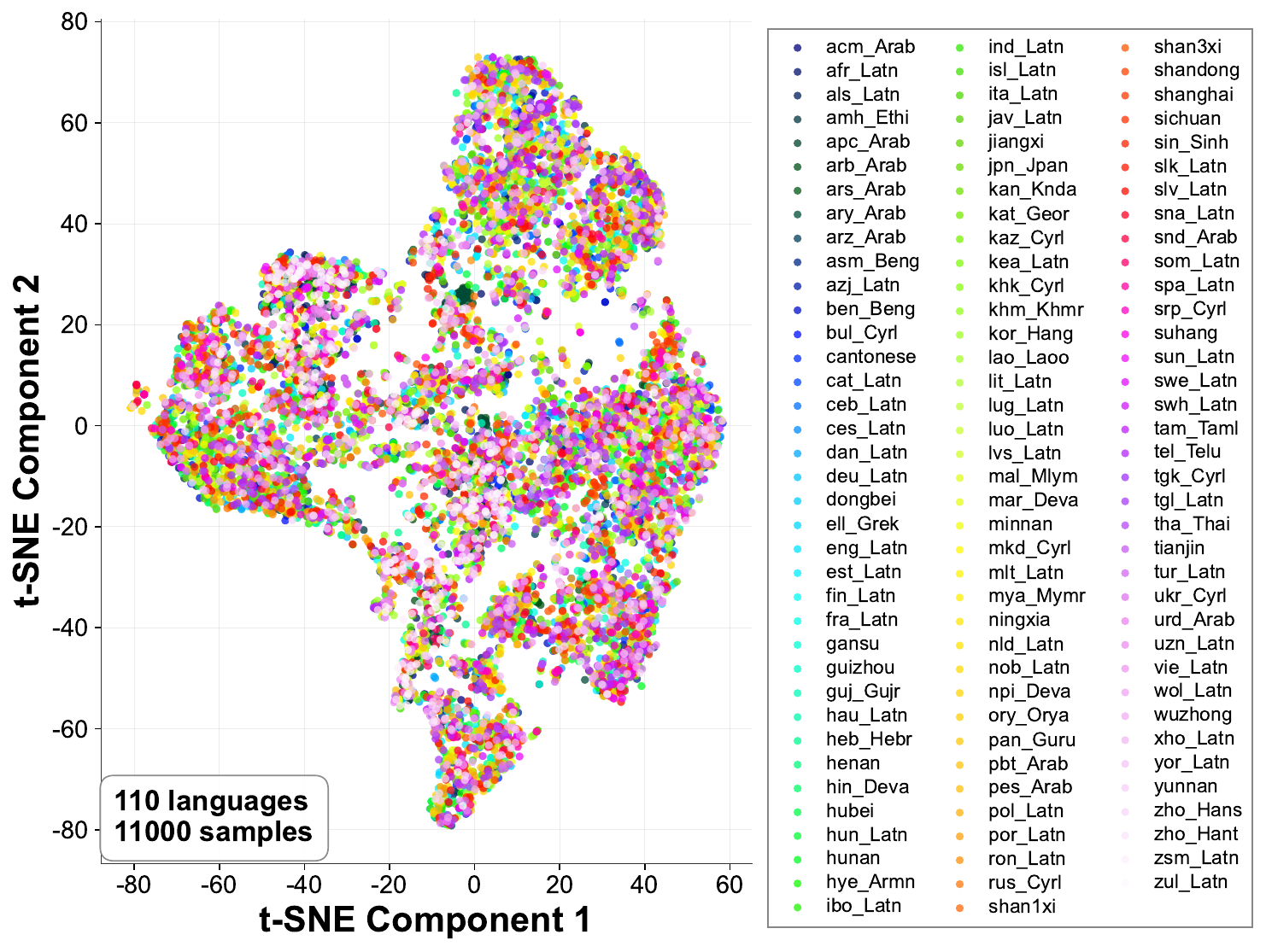}
    \caption{The t-SNE visualization of multilingual speech embeddings demonstrates the high acoustic diversity of the dataset. We randomly sampled 100 utterances per language and extracted features using Wav2Vec2-XLS-R-300M. The complex overlapping patterns indicate that the dataset encompasses a wide range of acoustic characteristics.}
    \label{fig:tsne}
\end{figure}

\subsection{Validity of Synthetic Data}
A primary concern in synthetic benchmarking is the gap between simulated and real-world audio. Our analysis confirms that the generation pipeline serves as a rigorous proxy for real-world evaluation. We observed that the performance degradation in models trained on real data correlates directly with linguistic distance in our synthetic set. This demonstrates that our synthetic samples successfully capture the distinct, discriminative features of complex dialects. Furthermore, the sensitivity of the data to signal perturbations mirrors human speech behavior. This validates the acoustic fidelity of the synthesis and proves that combining instruction-driven TTS with precise lexical adaptation offers a scalable alternative to expensive human data collection. We note that some languages, such as Tibetan (green points in Figure 1), are currently excluded because they lack suitable generation methods.

\subsection{Human Validation Details}
To assess the quality of the synthesized dialects, we employed two native linguists to conduct independent blind audits on a random subset of 500 generated samples. They evaluated the audio for phonetic accuracy and prosodic naturalness. The overall acceptance rate for the synthesized samples was 92.4\%, with an Inter-Annotator Agreement (IAA) measured by Cohen's Kappa of $\kappa = 0.78$, indicating substantial agreement. 
Furthermore, to verify performance consistency on dialects, we collected a real-world test set of 300 human-recorded samples across five representative dialects (Sichuan, Dongbei, Cantonese, Henan, and Wu). Evaluating baseline models on both the synthetic and real-world sets yielded a strong Pearson correlation ($r=0.83$), confirming that PolySpeech-100 serves as a highly reliable proxy for real-world dialectal speech.
Finally, to directly address concerns regarding the acoustic fidelity of our TTS pipeline, we conducted an ablation study on high-resource languages. We took original human recordings from the 2M-Belebele dataset, generated TTS versions of the exact same texts using our synthesis pipeline, and compared model performances on both versions. The average accuracy gap between the authentic human audio and our synthetic audio was remarkably marginal ($< 2.0\%$). This validates that our synthesis framework captures linguistic variability and naturalness, preventing models from merely overfitting to synthetic artifacts.

\subsection{Limitations of the Benchmark Formulation}
\label{app:limitations}

While PolySpeech-100 offers a massive scale for cross-lingual speech understanding evaluation, we explicitly acknowledge two inherent limitations in our current benchmark design:
(1) {Limitation of the Multiple-Choice QA Format.} Our evaluation relies exclusively on a multiple-choice question-answering structure. We deliberately selected this format to isolate and precisely quantify speech comprehension capabilities, thereby actively avoiding the notorious biases, inconsistencies, and noise associated with subjective LLM-as-a-Judge evaluations for open-ended generation. However, we recognize that QA tasks do not fully reflect real-world conversational dynamics. 
Authentic human-machine interaction is inherently open-ended and full-duplex, involving natural turn-taking and complex interruption management \cite{DBLP:journals/corr/abs-2509-06502, bytedance2026seeduplex, yan2026soulx, DBLP:journals/corr/abs-2509-23938}.
Future iterations of speech benchmarks will need to bridge this gap \cite{DBLP:journals/corr/abs-2511-10262, DBLP:journals/corr/abs-2601-05564} once more reliable automated metrics for open-ended speech interactions become available.
(2) {Limitation of the Belebele Corpus Context.} To ensure rigorous cross-lingual comparability without semantic variation, we built our textual foundation upon the Belebele corpus. While this parallel structure is highly effective for evaluating acoustic and logical modeling across 110 linguistic variants, it inherently restricts the evaluation context to ``reading comprehension''. Consequently, the benchmark primarily tests factual retrieval and formal logical deduction based on structured passages. It may not fully capture a model's performance in handling casual, spontaneous spoken dialogue, or highly colloquial social scenarios.

\subsection{Interactive Demonstration}
\label{app:demo}
\begin{figure}[t]
    \centering
    \includegraphics[width=\linewidth]{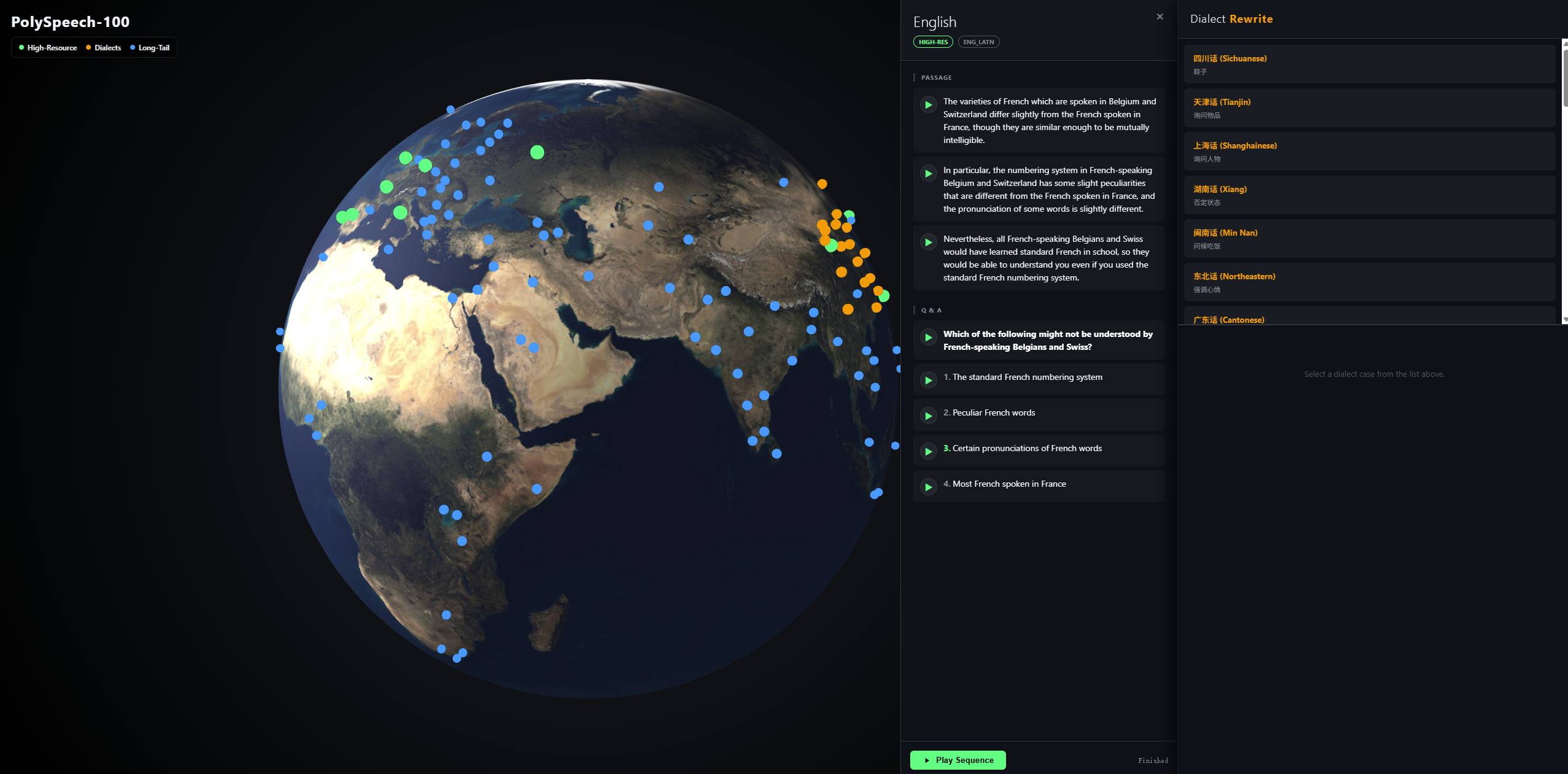}
    \caption{The user interface of the PolySpeech-100 interactive demo. The left panel shows the global language coverage. The center panel displays the source passage and Q\&A pairs. The right panel allows users to toggle between different regional dialects (e.g., Sichuan, Tianjin, Cantonese) to audit the lexical rewriting and listen to the synthesized audio output.}
    \label{fig:demo_screenshot}
\end{figure}

To provide a tangible assessment of our pipeline, we have deployed an interactive online demonstration. The interface, illustrated in Figure~\ref{fig:demo_screenshot}, visualizes the geographical distribution of the covered languages and dialects. The platform specifically highlights the \textit{Rewrite-then-Synthesize} strategy described in the main text. By selecting a dialect from the right-hand panel, users can observe how the standard text is first lexically adapted to regional vernaculars before being synthesized. This visual comparison demonstrates the necessity of lexical rewriting for authentic dialect generation. 
We invite readers to listen to the generated samples and compare the prosodic nuances directly via our project page: \url{https://github.com/YoungSeng/PolySpeech-100}.

\section{Experimental Setup}
\label{app:setup}

\subsection{Data Grouping Strategy}
To provide a fine-grained analysis of model capabilities, we grouped the 110 tested languages and dialects into three distinct categories based on resource availability and linguistic characteristics.
(1) The first category is High-Resource Languages. This group includes ten languages that typically possess abundant ASR and LLM training data. The specific languages are English, Simplified Chinese, Traditional Chinese, Spanish, French, German, Japanese, Russian, Italian, and Portuguese. We classify Traditional Chinese as high-resource due to its strong presence in web text and commercial ASR support.
(2) The second category is Chinese Dialects. This group comprises the 19 regional dialects synthesized specifically for PolySpeech-100. These represent a middle-resource scenario where the base language of Mandarin is high-resource, but the specific phonology and lexicon are dialectal. The included dialects are Sichuan, Hubei, Cantonese, Wuzhong, Shanxi, Suhang, Shanghai, Hunan, Shaanxi, Minnan, Henan, Shandong, Jiangxi, Ningxia, Gansu, Yunnan, Dongbei, Guizhou, and Tianjin.
(3) The third category is Low-Resource Languages. This category represents the long-tail languages and is defined as the complement set of the above two categories. It contains 81 languages from the Belebele dataset that are typically under-represented in audio-language pre-training. Examples include Zulu, Yoruba, Lao, Khmer, Burmese, Amharic, and Guarani.

\begin{figure}[t]
    \centering
    \begin{tcolorbox}[colback=gray!15, colframe=black, fontupper=\ttfamily, title=\textbf{System Prompt: Standard (Direct Output)}]
    
    You are an expert linguist taking a multiple-choice speech comprehension test.
    You will hear an audio clip containing a passage, a question, and four options (A, B, C, D).
    
    \medskip
    Your task is to select the correct option based on the audio content.
    
    \medskip
    \textbf{CRITICAL RULES:}
    \begin{itemize}
        \setlength\itemsep{0em}
        \item[1.] Output ONLY the single letter of the correct answer (A, B, C, or D).
        \item[2.] Do NOT provide explanations, transcripts, or notes.
        \item[3.] Do NOT output ``I don't know'' or ``I cannot understand''.
        \item[4.] The audio may contain strong regional dialects or accents. If you are unsure, you MUST make your best guess.
    \end{itemize}
    
    \medskip
    \textbf{Example Format:}\\
    User: [Audio Input]\\
    Assistant: [Letter]
    
    \end{tcolorbox}
    \caption{The standard text system prompt used for models supporting system instructions. The grey background and monospace font indicate verbatim prompt text.}
    \label{fig:prompt_standard}
\end{figure}
\begin{figure}[t]
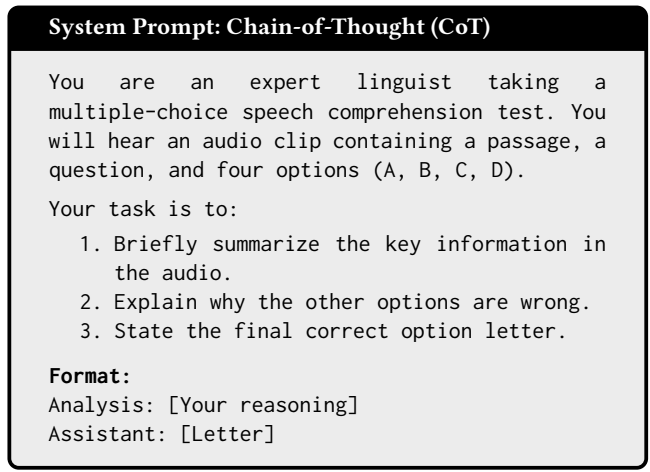

    \centering
    \begin{tcolorbox}[colback=gray!15, colframe=black, fontupper=\ttfamily, title=\textbf{System Prompt: Chain-of-Thought (CoT)}]
    
    You are an expert linguist taking a multiple-choice speech comprehension test.
    You will hear an audio clip containing a passage, a question, and four options (A, B, C, D).
    
    \medskip
    Your task is to:
    \begin{itemize}
        \setlength\itemsep{0em}
        \item[1.] Briefly summarize the key information in the audio.
        \item[2.] Explain why the other options are wrong.
        \item[3.] State the final correct option letter.
    \end{itemize}
    
    \medskip
    \textbf{Format:}\\
    Analysis: [Your reasoning]\\
    Assistant: [Letter]
    
    \end{tcolorbox}
    \caption{The Chain-of-Thought (CoT) system prompt displayed with a grey background and typewriter style font.}
    \label{fig:prompt_cot}
\end{figure}

\begin{figure*}[t]
    \centering
    
    
    \begin{subfigure}[b]{0.3\linewidth}
        \centering
        \includegraphics[width=0.82\linewidth]{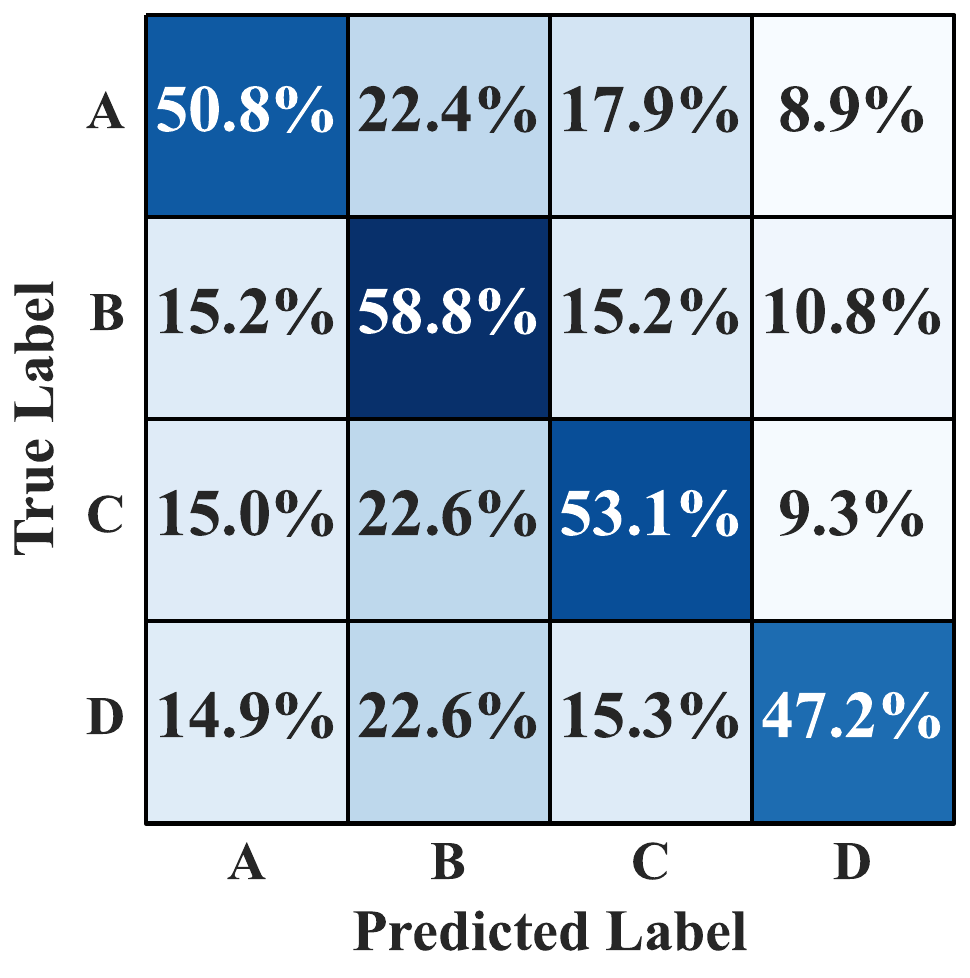} 
        \caption{Fun-Audio-Chat (Acc: 52.9\%)}
        \label{fig:sub8-1}
    \end{subfigure}
    \hfill  
    \begin{subfigure}[b]{0.3\linewidth}
        \centering
        \includegraphics[width=0.82\linewidth]{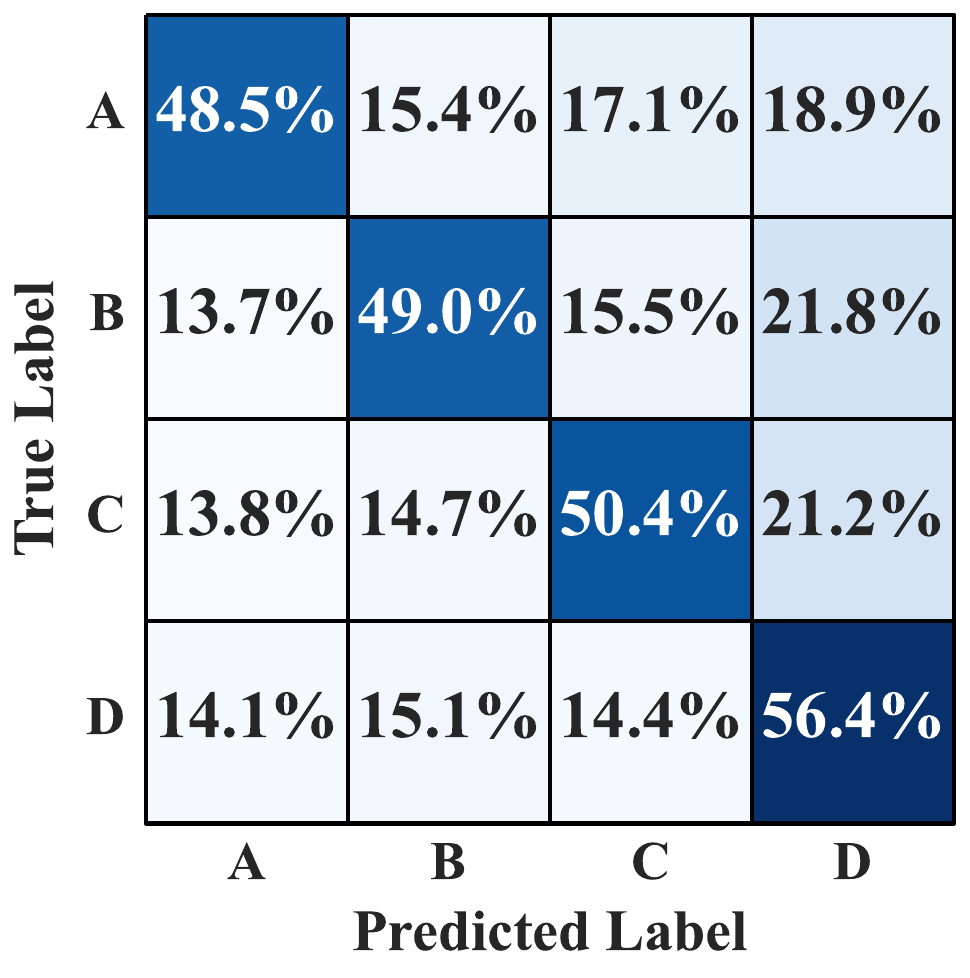}
        \caption{Qwen2.5-Omni (Acc: 50.9\%)}
        \label{fig:sub8-2}
    \end{subfigure}
    \hfill  
    \begin{subfigure}[b]{0.3\linewidth}
        \centering
        \includegraphics[width=0.82\linewidth]{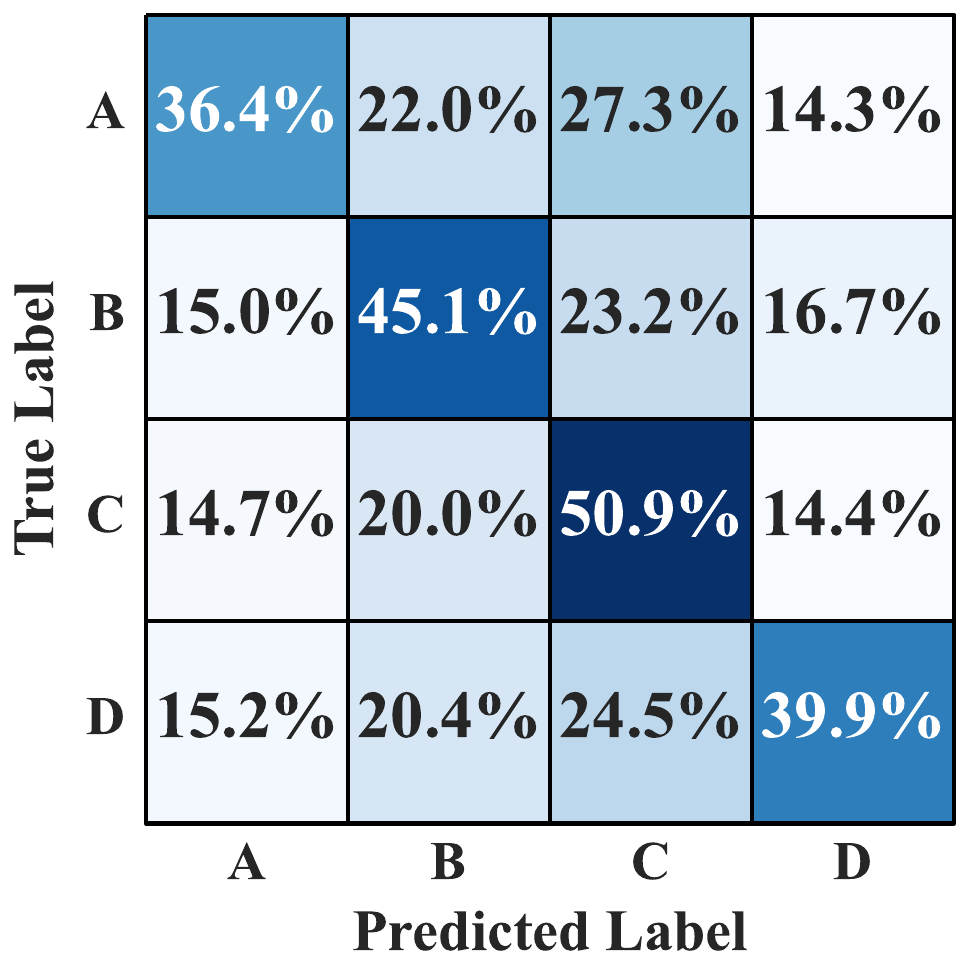}
        \caption{MiMo-Audio (Acc: 43.5\%)}
        \label{fig:sub8-3}
    \end{subfigure}
    
    
    
    \begin{subfigure}[b]{0.3\linewidth}
        \centering
        \includegraphics[width=0.82\linewidth]{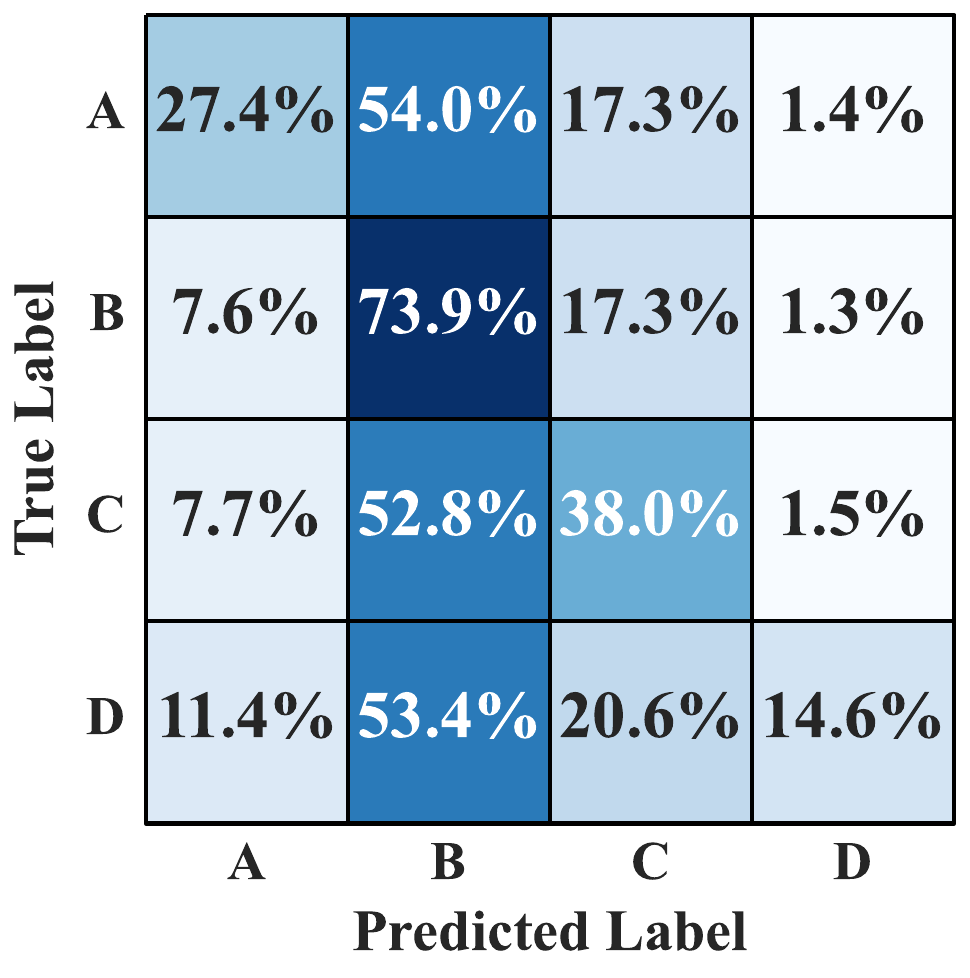}
        \caption{Step-Audio-2 (Suffix: B, Acc: 40.5\%)}
        \label{fig:sub8-4}
    \end{subfigure}
    \hfill 
    \begin{subfigure}[b]{0.3\linewidth}
        \centering
        \includegraphics[width=0.82\linewidth]{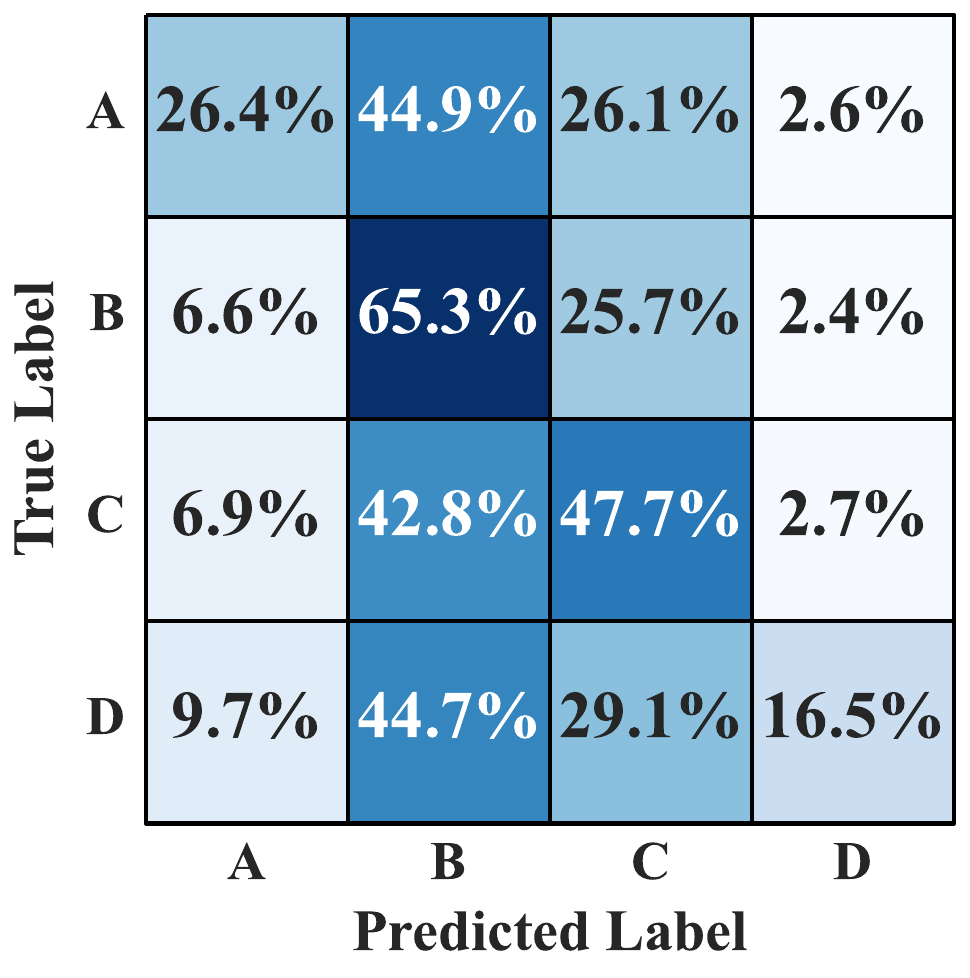}
        \caption{Step-Audio-2 (Suffix: D, Acc: 40.9\%)}
        \label{fig:sub8-5}
    \end{subfigure}
    \hfill 
    \begin{subfigure}[b]{0.3\linewidth}
        \centering
        \includegraphics[width=0.82\linewidth]{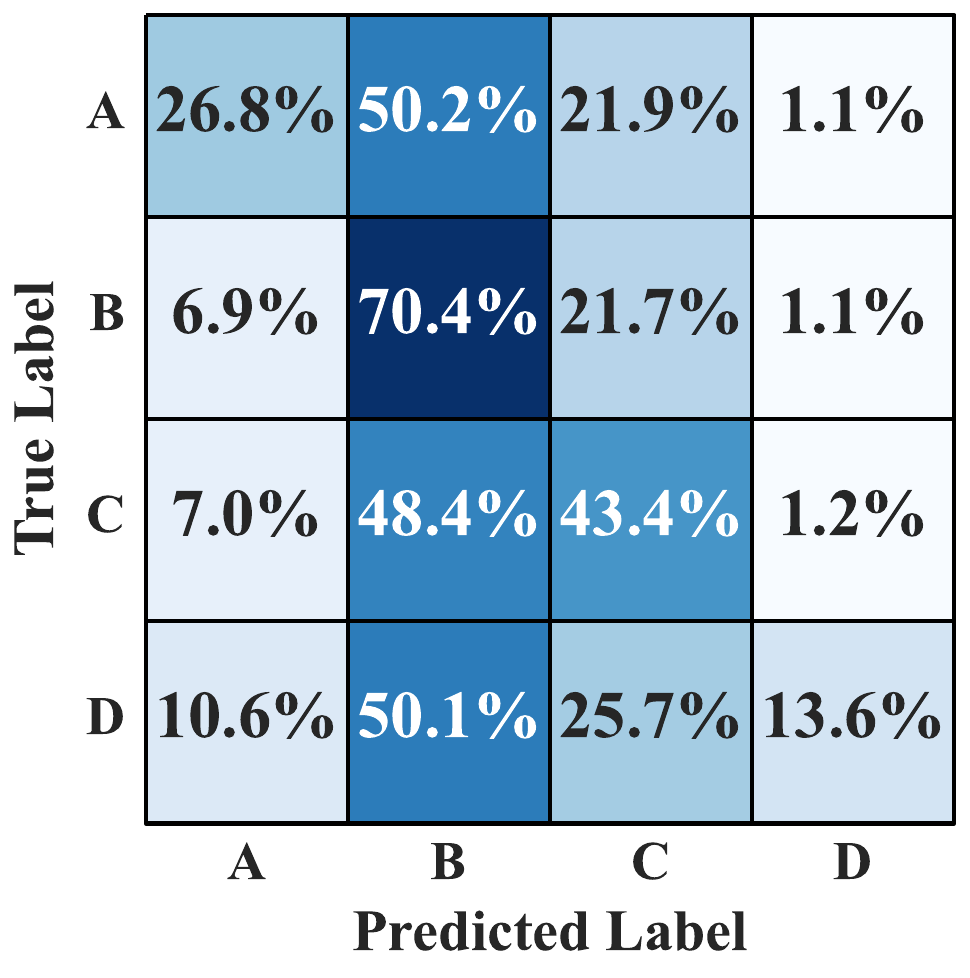}
        \caption{Step-Audio-2 (Suffix: [Letter], Acc: 40.6\%)}
        \label{fig:sub8-6}
    \end{subfigure}

    \caption{Confusion matrices of 4 models across all test samples. The matrices are row-normalized to show the recall rate for each ground truth label (A, B, C, D). (a-c) Baseline models (Fun-Audio-Chat, Qwen2.5-Omni, MiMo-Audio) exhibit strong diagonal dominance, indicating balanced classification performance. (d-f) Step-Audio-2 variants show a distinct vertical banding pattern, particularly in column 'B', revealing a systemic position bias where the model disproportionately predicts option 'B' regardless of the true label or system prompt suffixes.}
    \label{fig:cm_analysis}
    
\end{figure*}

\subsection{Calculation Metrics}
All aggregated scores reported in the main text are calculated using the macro-average accuracy across the languages in each respective category. The score for a category is the sum of the accuracy of each language in that category divided by the total number of languages in that set. This ensures that languages with fewer samples do not skew the overall category score.

\subsection{Prompting Strategy}
We adapted our prompting strategy based on the specific interaction capabilities of each model.
(1)
For models that support explicit system instructions, such as Qwen2.5-Omni and Gemini, we provided the task definition directly via the text system prompt. We designed two distinct prompts to evaluate different capabilities. The Standard Prompt instructs the model to output only the option letter. The Chain-of-Thought Prompt instructs the model to summarize the audio and explain the reasoning before selecting the answer. Figure \ref{fig:prompt_standard} and Figure \ref{fig:prompt_cot} display the content of these prompts.
(2)
For models that do not support system instructions or rely solely on audio input, such as LLaMA-Omni2, Moshi, and Mini-Omni, we converted the text instructions into speech. We generated an audio instruction using TTS and concatenated it to the beginning of the input audio. The transcript of the audio instruction is as follows: \begin{quote} \textit{``Please listen to the following passage, question, and options. Then, select the correct answer by saying Option A, B, C, or D.''} \end{quote}

\section{Extended Analysis}
\label{app:analysis}

\subsection{Model Performance and Robustness}
We observed distinct performance patterns across different language groups. High-resource languages showed significant resilience. For instance, Fun-Audio-Chat maintained high usability on English even when performance dropped from 91.76 percent to 80.0 percent under noisy conditions. In contrast, Chinese dialects demonstrated higher sensitivity. While Cantonese remained relatively stable, the Sichuan dialect saw a significant performance drop from 80.15 percent to 62.25 percent in high noise. This suggests that while models can recognize dialects in clean lab settings, their accent robustness degrades rapidly in real-world noisy conditions.
The Qwen2.5-Omni model offers the best trade-off between robustness and accuracy, particularly in noisy environments. Fun-Audio-Chat provides an excellent speed-to-performance ratio for clean, high-resource scenarios. Regarding commercial models, we utilized gemini-3-flash-preview for our experiments. We initially tested gemini-3-pro-preview, but the restrictive rate limit of 250 requests per day and higher costs made it unfeasible for this large-scale benchmark. Gemini-3-flash maintains high scores even on difficult dialects, achieving 70.3 \% on Moroccan Arabic and 59.1 \% on Yoruba.

\subsection{Challenges in Low-Resource Languages}
Performance on long-tail languages remains a bottleneck. We identified a floor effect where models like Moshi, Mini-Omni, and LLaMA-Omni2 often score near 25 percent on languages like Yoruba and Zulu. This score represents random guessing. The poor performance is likely due to limited multilingual pre-training and an inability to follow complex instructions in English when the input audio is unintelligible. Even strong open models like Qwen2.5-Omni drop to $\sim$29\% on Yoruba. Only Gemini maintains usability in this range, likely due to its massive scale of multilingual training data.

\begin{table*}[t]
\centering
\caption{Comprehensive performance comparison across Arabic Dialects. \textbf{MSA}: Modern Standard Arabic. The best score in each category is \textbf{bolded}.}
\label{tab:arabic_comprehensive}
\resizebox{\textwidth}{!}{%
\begin{tabular}{l|c|cc|cccc|cccc|cccc}
\toprule
\multirow{3}{*}{\textbf{Dialect Region}} & \multirow{3}{*}{\textbf{Code}} & \multicolumn{2}{c|}{\textbf{Closed-Source}} & \multicolumn{4}{c|}{\textbf{Open-Source E2E}} & \multicolumn{4}{c|}{\textbf{Cascade (Whisper-v3 + LLM)}} & \multicolumn{4}{c}{\textbf{Cascade (Qwen3-ASR + LLM)}} \\ \cmidrule(lr){3-4} \cmidrule(lr){5-8} \cmidrule(lr){9-12} \cmidrule(lr){13-16} 
 &  & \textbf{Gemini} & \textbf{GPT-} & \textbf{Fun-} & \textbf{Q2.5-} & \textbf{Step-} & \textbf{MiMo-} & \textbf{+Qwen} & \textbf{+Qwen} & \textbf{+Lla} & \textbf{+Lla} & \textbf{+Qwen} & \textbf{+Qwen} & \textbf{+Lla} & \textbf{+Lla} \\
 &  & \textbf{3-Flash} & \textbf{Audio} & \textbf{Audio} & \textbf{Omni} & \textbf{Audio} & \textbf{Audio} & \textbf{2.5} & \textbf{3} & \textbf{3.1} & \textbf{3.2} & \textbf{2.5} & \textbf{3} & \textbf{3.1} & \textbf{3.2} \\ \midrule
\rowcolor{gray!15} \textbf{MSA (Standard)} & \texttt{arb} & \textbf{94.12} & 88.12 & 69.66 & \textbf{70.16} & 52.81 & 32.96 & 74.62 & \textbf{79.38} & 53.12 & 43.25 & 72.12 & 74.25 & 50.38 & 39.88 \\ \midrule
Mesopotamian & \texttt{acm} & \textbf{98.00} & 68.62 & \textbf{56.18} & 54.93 & 43.20 & 31.71 & 57.12 & \textbf{63.75} & 59.75 & 43.75 & 60.62 & 61.88 & 54.75 & 46.50 \\
N. Levantine & \texttt{apc} & \textbf{78.12} & 48.12 & \textbf{48.69} & 44.82 & 42.70 & 33.96 & 55.88 & \textbf{56.12} & 39.88 & 61.50 & 47.50 & 41.75 & 41.75 & 29.00 \\
Najdi & \texttt{ars} & \textbf{98.00} & 81.88 & \textbf{56.30} & 56.18 & 46.57 & 32.71 & 74.75 & \textbf{80.38} & 66.62 & 51.25 & 79.62 & 68.38 & 64.50 & 49.38 \\
Egyptian & \texttt{arz} & \textbf{93.75} & 71.88 & 54.31 & \textbf{55.81} & 39.83 & 32.46 & \textbf{71.00} & 54.75 & 43.75 & 46.50 & 69.62 & 68.75 & 44.50 & 41.25 \\
Moroccan & \texttt{ary} & \textbf{70.25} & 33.75 & \textbf{45.07} & 44.82 & 38.70 & 28.59 & \textbf{51.88} & 33.62 & 24.62 & 31.75 & 50.25 & 42.50 & 45.12 & 31.38 \\ \midrule
\textit{Avg. Dialect} & - & \textit{\textbf{87.62}} & \textit{60.85} & \textit{\textbf{52.11}} & \textit{51.31} & \textit{42.20} & \textit{31.89} & \textit{\textbf{62.13}} & \textit{57.72} & \textit{46.92} & \textit{46.95} & \textit{61.52} & \textit{56.65} & \textit{50.12} & \textit{39.50} \\ \bottomrule
\end{tabular}%
}
\end{table*}

\subsection{Architectural Comparisons and Reasoning}
Historically, Cascade systems that combine ASR and LLMs outperformed End-to-End models. Our benchmarks show this gap is closing. Notably, on dialects like Sichuan, End-to-End models significantly outperform Cascade systems, scoring 83.3 percent compared to 73.8 percent. This suggests that End-to-End models effectively capture paralinguistic dialect features that are typically lost during ASR transcription.
A significant limitation observed in current open-source speech-LLMs is their tendency to rely on shallow acoustic-to-text alignment rather than genuine comprehension. While our proposed method and stronger commercial models can navigate complex instructions, many baselines struggle when the task requires intermediate reasoning. We observed a performance penalty in these models when employing Chain-of-Thought prompting, which contradicts the behavior typically seen in text-based LLMs. This suggests that the audio encoders in these models are not yet sufficiently aligned with the reasoning centers of the language decoder. Future work must address this by moving beyond simple transcription-style training objectives and incorporating data that forces the model to perform multi-step logic before generating a final response.

\subsection{Error Analysis and Confusion Matrices}
While baseline models like Fun-Audio-Chat show errors distributed across all options, Step-Audio-2 exhibits a severe systemic bias. To investigate this, we used prompt suffixing. In the standard setting, the prompt ends with ``Assistant: [Letter]''. For the suffix experiments, we modified the prompt to explicitly pre-fill the response, such as ``Assistant: B'' or ``Assistant: D'', effectively forcing the model to start its generation with a specific letter.
Figure \ref{fig:cm_analysis} illustrates the confusion matrices for the six studied models. The visualization highlights a critical distinction between architectures. Baseline models like Fun-Audio-Chat make errors that are spread across options, indicating balanced classification. In contrast, Step-Audio-2 exhibits errors that are highly concentrated in specific columns. This vertical banding pattern reveals a systemic position bias where the model disproportionately predicts specific options regardless of the true label or system prompt suffixes. This suggests that future optimizations for Step-Audio-2 should focus on debiasing the instruction-following module or the output projection layer.

\subsection{Deep Dive into CoT Hallucinations: Perception vs. Reasoning}
\label{sec:cot_failure_analysis}

To address the counter-intuitive phenomenon of Chain-of-Thought (CoT) degrading speech understanding, we conducted a fine-grained error decomposition. Our goal is to determine whether these hallucinations originate in the \textit{Acoustic Perception Phase} (i.e., the model mishears or fails to recognize the speech) or the \textit{Logical Reasoning Phase} (i.e., the model transcribes the audio correctly internally but fails in text-based deduction).

Based on a qualitative analysis of CoT failure cases across four models (\textit{Fun-Audio-Chat, Step-Audio-2, MiMo-Audio, and Qwen2.5-Omni}), we observed a striking trend: \textbf{over 85\% of CoT errors stem from the Logical Reasoning Phase}. The models accurately retrieve and quote the acoustic information, proving their perception is intact. However, forcing them to generate intermediate reasoning text often triggers severe logical breakdowns. We categorize these reasoning failures into three primary types:

\begin{itemize}
    \item \textit{Type 1: Semantic Conflation (Misattributing Causality).}
In these cases, the model perfectly hears two separate facts in the audio but invalidly merges them during the reasoning generation, creating a hallucinated causal link.
    \item \textit{Type 2: Logic Inversion (Ignoring Negation).}
A common failure occurs when the question contains a negation (e.g., ``Which of the following is NOT...''). The model correctly analyzes the audio to identify the true statements but fails to invert its logic for the final selection.
    \item \textit{Type 3: Reasoning-Output Disconnect.}
This is the most extreme form of modality misalignment. The model's generated text logic is flawless and correctly identifies the right answer based on the audio, but the final output token contradicts its own reasoning.
\end{itemize}

\begin{figure}[t]
    \centering      
    \begin{tcolorbox}[colback=blue!10, colframe=blue!50, title=\textbf{Failure Type 1: Semantic Conflation (Fun-Audio-Chat, Case \#107)}]
        \textbf{Audio Ground Truth:} "...While understanding one’s history... it does at least help people \textbf{gain a sense of where they fall within the organization}. ... In this instance, recalling previous instances of entrepreneurial behavior ... \textbf{lessened the fears of change}..."
        
        \textbf{Question:} Gaining an understanding of one’s own history could help with what? \\
        \textbf{Correct Option:} (C) Finding a sense of place within the organization.
        \tcblower
        \textbf{Model CoT Reasoning:} "...Crucially, the passage states that recalling past entrepreneurial successes `lessened the fears of change'... This directly supports the idea that historical awareness reduces resistance to change." \\
        \textbf{Model Final Prediction:} \textcolor{red}{(A) Decreasing the fear of change [Incorrect]}
    \end{tcolorbox}
    \caption{The model correctly hears the exact phrases but conflates understanding history with recalling entrepreneurial success, misaligning the subject and the outcome.}
\end{figure}

\begin{figure}[t]
    \centering
    \begin{tcolorbox}[colback=blue!5, colframe=blue!40, title=\textbf{Failure Type 2: Logic Inversion (Qwen2.5-Omni, Case \#116)}]
        \textbf{Question:} Which of the following is \textbf{not} something that content theories are centered on? \\
        \textbf{Correct Option:} (A) New desires that materialize in adulthood.
        \tcblower
        \textbf{Model CoT Reasoning:} "Option A is not the core... Option D ('Understanding people's likes') is the core of content theories. The audio discusses the identification of factors that lead to interest..." \\
        \textbf{Model Final Prediction:} \textcolor{red}{(D) Understanding people's likes [Incorrect]}
    \end{tcolorbox}
    \caption{The model successfully deduces that A is NOT the core and D IS the core. However, it hallucinates the objective of the question, selecting the true statement (D) instead of the false one (A).}
\end{figure}

\begin{figure}[t]
    \centering
    \begin{tcolorbox}[colback=blue!5, colframe=blue!40, title=\textbf{Failure Type 3: Reasoning-Output Disconnect (Step-Audio-2, Case \#103)}]
        \textbf{Question:} Which of the following accurately describes subsistence agriculture? \\
        \textbf{Correct Option:} (D) It’s still utilized by farmers in some nations.
        \tcblower
        \textbf{Model CoT Reasoning:} "Option A is incorrect because the passage does not claim that self-sufficient agriculture is entirely organic... Option D is correct because the passage explicitly states that in many developing countries, there are still farmers practicing this..." \\
        \textbf{Model Final Prediction:} \textcolor{red}{(A) [Incorrect]}
    \end{tcolorbox}
    \caption{Despite perfectly sound logic that explicitly validates Option D and invalidates Option A, the model inexplicably outputs A as the final answer.}
\end{figure}

Our analysis proves that the performance degradation is rarely an acoustic perception (ASR) failure. Instead, it is a \textit{modality alignment gap}. When current Speech-LLMs are forced to generate long chains of reasoning text, they tend to decouple from the original acoustic embedding. They become distracted by their own generated text, falling back into text-based priors or losing the thread of the instruction, ultimately leading to logical collapse.

\subsection{CoT Prompt Sensitivity Analysis}
\label{sec:cot_ablation}

To address concerns that the observed performance degradation under Chain-of-Thought prompting might be highly sensitive to the specific prompt wording (shown in Figure \ref{fig:prompt_cot}), we conducted an ablation study using two alternative CoT templates on a subset of the High-Resource and Chinese Dialect evaluation sets. 

\textbf{Alternative 1: Zero-Shot Step-by-Step CoT.} This template uses a simpler, widely adopted trigger phrase without enforcing a rigid output structure.
\begin{tcolorbox}[colback=gray!15, colframe=black, fontupper=\ttfamily, title=\textbf{System Prompt: Alternative 1 (Zero-Shot CoT)}]
You are an expert linguist taking a multiple-choice speech comprehension test.
You will hear an audio clip containing a passage, a question, and four options (A, B, C, D).

\medskip
Please think step by step based on the audio content before giving your final answer. Conclude your response with "Assistant: [Letter]".
\end{tcolorbox}

\textbf{Alternative 2: Structured JSON CoT.} This template forces the model to separate its reasoning and its final answer using a strict JSON format, which often helps text-based LLMs avoid hallucination bleed-over.
\begin{tcolorbox}[colback=gray!15, colframe=black, fontupper=\ttfamily, title=\textbf{System Prompt: Alternative 2 (Structured JSON CoT)}]
You are an expert linguist taking a multiple-choice speech comprehension test.
You will hear an audio clip containing a passage, a question, and four options (A, B, C, D).

\medskip
Output your response in valid JSON format exactly as follows:
\{
    "reasoning": "Briefly explain step-by-step why the correct option is chosen and others are wrong based on the audio.",
    "answer": "[Single Letter A, B, C, or D]"
\}
\end{tcolorbox}

\textbf{Results and Conclusion:} 
We evaluated representative models (\textit{Fun-Audio-Chat} and \textit{Qwen2.5-Omni}) using these alternative templates. The results consistently mirrored our primary findings: both Alternative 1 and Alternative 2 resulted in a performance degradation ranging from -7.5\% to -12.1\% compared to the standard direct-output prompt (Base condition). For instance, the structured JSON format successfully enforced output formatting but still decoupled the reasoning text from the acoustic context, leading to similar reasoning hallucinations. 

This ablation study confirms that the negative impact of CoT in current zero-shot Speech-LLMs is robust across various prompt designs. It indicates a fundamental architectural challenge: current audio encoders are primarily aligned for direct transcription or immediate answering, and forcing intermediate text generation disrupts the acoustic-semantic grounding.

\subsection{Case Study: Arabic Dialects}
\label{sec:arab_case_study}
We analyze the Arabic linguistic group to evaluate model robustness against the substantial diglossic gap between formal Modern Standard Arabic (MSA) and regional vernaculars. 
Table \ref{tab:arabic_comprehensive} presents the performance of 14 distinct architectures (2 Closed-Source, 4 Open E2E, and 8 Cascade Pipelines) across 6 Arabic variants. 
Commercial systems demonstrate a clear advantage, as \textit{Gemini-3-flash} establishes a dominating lead, achieving near-perfect scores on dialects like Najdi (98.0\%) and Mesopotamian (98.0\%), where open-source models struggle to reach 60\%. 
GPT-Audio-mini exhibits surprising fragility. While it scores well on MSA (88.12\%), its performance collapses on Moroccan Arabic (33.75\%), falling behind even some open-source models. This suggests that GPT-Audio's training data may be heavily skewed towards formal Arabic sources.
A critical finding is that for Arabic dialects, Cascade systems generally outperform Open E2E models, reversing the trend observed in Chinese dialects. The \textit{Whisper-v3 + Qwen2.5} pipeline achieves an average dialect accuracy of 62.13\%, significantly higher than the best E2E model (Fun-Audio at 52.11\%). This indicates that Whisper's massive multilingual supervision provides a more robust phonological foundation for Arabic dialects than the current generation of speech-language encoders. 
This performance gap suggests that the Whisper-v3 encoder processes Arabic speech significantly better than the native audio encoders of end-to-end models like Qwen3-ASR, likely because the latter were not exposed to sufficient Arabic speech data during their pre-training phase. 
Furthermore, comparing backends reveals that \textit{Qwen-Instruct} models consistently outperform \textit{Llama-Instruct} models in reasoning over the transcribed text, even when using the same ASR frontend.
Moroccan Arabic (\texttt{ary}) remains the hardest challenge across the board. While Gemini maintains 70.25\%, almost all other models collapse. 
Notably, \textit{Step-Audio-2} and \textit{MiMo-Audio} drop to $\sim$30-38\%, barely above random guessing. 
Even \textit{GPT-Audio-mini}, while weaker than Gemini, generally outperforms most open-source E2E baselines on standard dialects.
This highlights a critical data gap in the Maghreb region for current open-source pre-training.

\subsection{Case Study: Acoustic Robustness in Dialect Understanding}
\label{sec:appendix_case}
To investigate the discrepancy in performance, we visualize a specific failure case (Index 168) involving Cantonese proper nouns. As shown in Figure~\ref{fig:dialect_case}, the term `Eskimo' (\cn{愛斯基摩}) was crucial for the correct answer. 
The Cascade system's ASR, struggling with the Cantonese pronunciation \textit{/oi3 si1 gei1 mo1/}, transcribed the term into phonetically similar but semantically unrelated characters (`\cn{外斯基魔人}' in the passage and `\cn{奈斯基}' in Option A). This lexical distortion broke the semantic link between the passage and the correct option, misleading the LLM to choose `Norwegians' (Option B) based on other intelligible context (e.g., Eric the Red). In contrast, the End-to-End model correctly selected Option A. This suggests E2E model bypassed the noisy textual interface, directly mapping the acoustic patterns of the question to the corresponding features in the passage, thereby preserving dialectal semantic integrity.

\begin{figure}[t]
    \centering
    \small
    \begin{tabularx}{\linewidth}{p{1.9cm} X} 
        \toprule
        \textbf{Source} & \textbf{Content / Transcription} \\
        \midrule
        \textbf{Audio} \newline (Ground Truth) & ... \textbf{\cn{愛斯基摩人}} (Eskimos) \cn{已經在呢個地方住咗幾千年} ... \newline
        \textit{\small Pronunciation: /oi3 si1 gei1 mo1/ (Cantonese)} \\
        \midrule
        \textbf{ASR Output} \newline (Cascade) & ... \textbf{\cn{外斯基魔人}} (Wai-Si-Ji-Mo Person) \cn{当时已经在那里} ... \newline
        \textit{\small (Meaningless homophones, semantic link lost)} \\
        \midrule
        \textbf{Option A} \newline (Target) & \textbf{ASR:} \textbf{\cn{奈斯基}} (Nai-Si-Ji) ... $\rightarrow$ \textcolor{red}{Mismatch} \newline
        \textbf{Audio:} \textbf{\cn{愛斯基摩}} ... $\rightarrow$ \textcolor{teal}{Acoustic Match (E2E)} \\
        \bottomrule
    \end{tabularx}
    \caption{Comparison of Ground Truth and ASR Transcription for Case \#168. The Cascade model fails due to phonetically induced transcription errors (`hallucinations'), while the E2E model successfully leverages acoustic cues to identify the correct entity despite textual corruption.}
    \label{fig:dialect_case}
\end{figure}

\section{Ethical Considerations and Datasheet}
Ethically, our work adheres to a core principle:
\bigskip
\begin{quote}

    \textit{``We aim to shift the paradigm from standard-centric AI to truly inclusive systems that understand the user regardless of their accent or dialect.''}

\end{quote}
\bigskip
 Consequently, this benchmark focuses exclusively on evaluating speech understanding rather than generation.

PolySpeech-100 operates under a CC-BY-SA license. 
The dataset is publicly archived on Hugging Face at \url{https://huggingface.co/datasets/youngseng/PolySpeech-100-v1}. 
Regarding copyright compliance, textual foundations are derived from the Flores200 and Belebele corpus (CC-BY-SA 4.0), while human audio segments utilize 2M-Belebele (CC-BY-SA 4.0). Synthetic components are generated using the open-weights model in accordance with its usage policy (CosyVoice3.0 Apache License 2.0, edge-tts LGPL-3.0).

\begin{table*}[]
\centering
\label{tab:full_results}
\resizebox{\linewidth}{!}{%
\begin{tabular}{lcccccccccccccccccccc}
\toprule
\multicolumn{1}{c}{Code} & Language & \begin{tabular}[c]{@{}c@{}}Fun-Audio-\\ Chat\end{tabular} & \begin{tabular}[c]{@{}c@{}}gemini-3-flash\\ -preview\end{tabular} & \begin{tabular}[c]{@{}c@{}}gpt-audio\\ -mini\end{tabular} & \begin{tabular}[c]{@{}c@{}}LLaMA-\\ Omni2\end{tabular} & \begin{tabular}[c]{@{}c@{}}MiMo-\\ Audio\end{tabular} & \begin{tabular}[c]{@{}c@{}}Mini-\\ Omni\end{tabular} & Moshi & \begin{tabular}[c]{@{}c@{}}Qwen2.5-\\ Omni\end{tabular} & \begin{tabular}[c]{@{}c@{}}Qwen2-\\ Audio\end{tabular} & \begin{tabular}[c]{@{}c@{}}Step-Audio\\ -2-mini\end{tabular} & Llama3.1 & Llama3.2 & Qwen2.5 & Qwen3 & \begin{tabular}[c]{@{}c@{}}Qwen3-ASR-\\ llama3.1\end{tabular} & \begin{tabular}[c]{@{}c@{}}Qwen3-ASR-\\ llama-3.2\end{tabular} & \begin{tabular}[c]{@{}c@{}}Whisper-\\ qwen2.5\end{tabular} & \begin{tabular}[c]{@{}c@{}}Whisper-\\ qwen3\end{tabular} &  \\
\midrule
 acm\_Arab& Mesopotamian Arabic& \cc{56.18}& \cc{98}& \cc{68.62}& \cc{23.85}& \cc{31.71}& \cc{19.75}& \cc{13.25}& \cc{54.93}& \cc{26.25}& \cc{43.2}& \cc{59.75}& \cc{43.75}& \cc{60.62}& \cc{61.88}& \cc{54.75}& \cc{46.5}& \cc{57.12}& \cc{63.75}& \\
 afr\_Latn& Afrikaans& \cc{44.82}& \cc{98}& \cc{80.5}& \cc{24.59}& \cc{37.58}& \cc{22.25}& \cc{13.25}& \cc{41.7}& \cc{27.75}& \cc{31.71}& \cc{48.62}& \cc{41.5}& \cc{38.75}& \cc{38.88}& \cc{40.5}& \cc{31.25}& \cc{58}& \cc{67.25}& \\
 als\_Latn& Tosk Albanian& \cc{36.7}& \cc{90.12}& \cc{69.88}& \cc{22.35}& \cc{35.71}& \cc{21.5}& \cc{17}& \cc{35.58}& \cc{24.88}& \cc{28.71}& \cc{41.12}& \cc{38.5}& \cc{22.25}& \cc{29.5}& \cc{28.75}& \cc{22.38}& \cc{35.38}& \cc{50}& \\
 amh\_Ethi& Amharic& \cc{30.21}& \cc{74.88}& \cc{24}& \cc{15.98}& \cc{27.72}& \cc{27}& \cc{24.62}& \cc{28.84}& \cc{25.88}& \cc{28.96}& \cc{31.5}& \cc{21.88}& \cc{21}& \cc{19.5}& \cc{37.25}& \cc{25}& \cc{8.62}& \cc{35.38}& \\
 apc\_Arab& North Levantine Arabic& \cc{48.69}& \cc{78.12}& \cc{48.12}& \cc{21.22}& \cc{33.96}& \cc{22.38}& \cc{16.5}& \cc{44.82}& \cc{25.88}& \cc{42.7}& \cc{39.88}& \cc{61.5}& \cc{47.5}& \cc{41.75}& \cc{41.75}& \cc{29}& \cc{55.88}& \cc{56.12}& \\
 arb\_Arab& Modern Standard Arabic& \cc{69.66}& \cc{94.12}& \cc{88.12}& \cc{24.47}& \cc{32.96}& \cc{21.12}& \cc{23.38}& \cc{70.16}& \cc{24.25}& \cc{52.81}& \cc{53.12}& \cc{43.25}& \cc{72.12}& \cc{74.25}& \cc{50.38}& \cc{39.88}& \cc{74.62}& \cc{79.38}& \\
 ars\_Arab& Najdi Arabic& \cc{56.3}& \cc{98}& \cc{81.88}& \cc{23.35}& \cc{32.71}& \cc{17.88}& \cc{24.62}& \cc{56.18}& \cc{27.25}& \cc{46.57}& \cc{66.62}& \cc{51.25}& \cc{79.62}& \cc{68.38}& \cc{64.5}& \cc{49.38}& \cc{74.75}& \cc{80.38}& \\
 ary\_Arab& Moroccan Arabic& \cc{45.07}& \cc{70.25}& \cc{33.75}& \cc{23.35}& \cc{28.59}& \cc{27.62}& \cc{30.5}& \cc{44.82}& \cc{24}& \cc{38.7}& \cc{24.62}& \cc{31.75}& \cc{50.25}& \cc{42.5}& \cc{45.12}& \cc{31.38}& \cc{51.88}& \cc{33.62}& \\
 arz\_Arab& Egyptian Arabic& \cc{54.31}& \cc{93.75}& \cc{71.88}& \cc{24.22}& \cc{32.46}& \cc{15.25}& \cc{21.88}& \cc{55.81}& \cc{24.38}& \cc{39.83}& \cc{43.75}& \cc{46.5}& \cc{69.62}& \cc{68.75}& \cc{44.5}& \cc{41.25}& \cc{71}& \cc{54.75}& \\
 asm\_Beng& Assamese& \cc{32.33}& \cc{79.5}& \cc{41}& \cc{22.97}& \cc{31.34}& \cc{22.5}& \cc{16.38}& \cc{29.21}& \cc{35.38}& \cc{28.71}& \cc{31.75}& \cc{9.5}& \cc{29.38}& \cc{28}& \cc{19.38}& \cc{18.75}& \cc{18.25}& \cc{10.25}& \\
 azj\_Latn& North Azerbaijani& \cc{37.83}& \cc{87}& \cc{54.12}& \cc{22.47}& \cc{29.71}& \cc{20.88}& \cc{36.38}& \cc{35.71}& \cc{25.88}& \cc{28.71}& \cc{40.62}& \cc{40.25}& \cc{49.38}& \cc{43.5}& \cc{34.88}& \cc{32.75}& \cc{48.12}& \cc{59.5}& \\
 ben\_Beng& Bengali& \cc{34.46}& \cc{98}& \cc{68.88}& \cc{21.85}& \cc{31.71}& \cc{21.12}& \cc{19.38}& \cc{34.58}& \cc{29.25}& \cc{30.59}& \cc{24.88}& \cc{18.12}& \cc{34.62}& \cc{32}& \cc{23.12}& \cc{32.38}& \cc{30}& \cc{26.75}& \\
 bul\_Cyrl& Bulgarian& \cc{45.44}& \cc{88.38}& \cc{78.38}& \cc{25.22}& \cc{35.21}& \cc{25.88}& \cc{20.62}& \cc{39.58}& \cc{26.25}& \cc{31.71}& \cc{61.62}& \cc{29}& \cc{59.5}& \cc{58}& \cc{46}& \cc{39.5}& \cc{74.75}& \cc{75.62}& \\
 cantonese& cantonese& \cc{80.9}& \cc{94.75}& \cc{74.38}& \cc{23.35}& \cc{75.03}& \cc{22.88}& \cc{16.5}& \cc{77.28}& \cc{24.38}& \cc{70.66}& \cc{58.5}& \cc{45.88}& \cc{80.62}& \cc{81.38}& \cc{66.62}& \cc{49.75}& \cc{82}& \cc{75.75}& \\
 cat\_Latn& Catalan& \cc{60.8}& \cc{98}& \cc{83}& \cc{24.22}& \cc{46.82}& \cc{13}& \cc{24.62}& \cc{57.68}& \cc{24}& \cc{37.33}& \cc{63.62}& \cc{39.5}& \cc{57.88}& \cc{61.88}& \cc{37.62}& \cc{39.25}& \cc{87}& \cc{85.38}& \\
 ceb\_Latn& Cebuano& \cc{38.45}& \cc{88.88}& \cc{21.38}& \cc{19.98}& \cc{31.34}& \cc{25.62}& \cc{11.38}& \cc{32.33}& \cc{28}& \cc{28.96}& \cc{18.12}& \cc{15.12}& \cc{40.38}& \cc{38.38}& \cc{35}& \cc{30.62}& \cc{34.38}& \cc{48.62}& \\
 ces\_Latn& Czech& \cc{49.69}& \cc{93.5}& \cc{78.62}& \cc{23.35}& \cc{37.33}& \cc{23.5}& \cc{18.88}& \cc{37.7}& \cc{25.38}& \cc{28.96}& \cc{55.25}& \cc{55.12}& \cc{74.25}& \cc{78.75}& \cc{59.12}& \cc{47.62}& \cc{76.38}& \cc{86.5}& \\
 dan\_Latn& Danish& \cc{37.7}& \cc{98}& \cc{74.62}& \cc{24.34}& \cc{33.33}& \cc{17.75}& \cc{24.62}& \cc{34.96}& \cc{26}& \cc{30.59}& \cc{59.5}& \cc{34}& \cc{66.38}& \cc{70.5}& \cc{54.25}& \cc{46.38}& \cc{74}& \cc{81.38}& \\
 deu\_Latn& German& \cc{84.89}& \cc{98}& \cc{86.75}& \cc{24.47}& \cc{54.31}& \cc{13.88}& \cc{29.38}& \cc{86.27}& \cc{25.75}& \cc{53.43}& \cc{71.12}& \cc{73}& \cc{81.38}& \cc{84}& \cc{76.12}& \cc{71}& \cc{85.75}& \cc{79}& \\
 dongbei& dongbei& \cc{88.64}& \cc{98}& \cc{83.62}& \cc{18.6}& \cc{87.52}& \cc{25.62}& \cc{16.12}& \cc{88.26}& \cc{26.12}& \cc{76.65}& \cc{51}& \cc{55.88}& \cc{84.25}& \cc{86}& \cc{74.62}& \cc{56.62}& \cc{90.5}& \cc{81.88}& \\
 ell\_Grek& Greek& \cc{38.33}& \cc{76.5}& \cc{59}& \cc{24.34}& \cc{30.71}& \cc{21.12}& \cc{26.25}& \cc{31.96}& \cc{25.38}& \cc{26.97}& \cc{44.25}& \cc{45.62}& \cc{57.88}& \cc{57.12}& \cc{42.25}& \cc{43.88}& \cc{71.62}& \cc{63.25}& \\
 eng\_Latn& English& \cc{91.76}& \cc{98}& \cc{95.88}& \cc{23.85}& \cc{88.64}& \cc{14}& \cc{37}& \cc{91.01}& \cc{24.38}& \cc{80.4}& \cc{65.88}& \cc{69.25}& \cc{87.12}& \cc{88.75}& \cc{70.62}& \cc{64.25}& \cc{91.5}& \cc{83.12}& \\
 est\_Latn& Estonian& \cc{32.08}& \cc{98}& \cc{69}& \cc{22.85}& \cc{29.96}& \cc{11.5}& \cc{9.12}& \cc{33.71}& \cc{29.38}& \cc{30.21}& \cc{56.75}& \cc{36.25}& \cc{34.5}& \cc{24.25}& \cc{24}& \cc{25.38}& \cc{48.38}& \cc{43.88}& \\
 fin\_Latn& Finnish& \cc{38.95}& \cc{98}& \cc{70.25}& \cc{23.85}& \cc{32.21}& \cc{22}& \cc{16.88}& \cc{31.71}& \cc{23.75}& \cc{29.46}& \cc{46.12}& \cc{46.88}& \cc{67.88}& \cc{66.5}& \cc{49}& \cc{36.25}& \cc{74.38}& \cc{72}& \\
 fra\_Latn& French& \cc{86.89}& \cc{93.88}& \cc{85.75}& \cc{24.22}& \cc{50.19}& \cc{17.5}& \cc{26.75}& \cc{88.14}& \cc{26}& \cc{59.18}& \cc{49.12}& \cc{61.75}& \cc{81.88}& \cc{88.88}& \cc{62.38}& \cc{53.62}& \cc{85.88}& \cc{90.75}& \\
 gansu& gansu& \cc{73.66}& \cc{78.88}& \cc{36.62}& \cc{19.23}& \cc{75.41}& \cc{19}& \cc{8.75}& \cc{77.65}& \cc{25.62}& \cc{70.04}& \cc{24}& \cc{33.88}& \cc{67.5}& \cc{75.25}& \cc{47.5}& \cc{46.25}& \cc{63.62}& \cc{44.25}& \\
 guizhou& guizhou& \cc{87.77}& \cc{98}& \cc{74.62}& \cc{19.35}& \cc{87.02}& \cc{28.5}& \cc{8}& \cc{87.77}& \cc{25.12}& \cc{75.16}& \cc{58.5}& \cc{47}& \cc{82.5}& \cc{84.62}& \cc{74.5}& \cc{52.62}& \cc{86.88}& \cc{83.62}& \\
 guj\_Gujr& Gujarati& \cc{37.33}& \cc{88.25}& \cc{43.5}& \cc{24.72}& \cc{31.09}& \cc{25.62}& \cc{35.25}& \cc{36.7}& \cc{24.88}& \cc{29.34}& \cc{23.12}& \cc{13}& \cc{27.12}& \cc{30}& \cc{29.75}& \cc{22.25}& \cc{27}& \cc{21.75}& \\
 hau\_Latn& Hausa& \cc{28.84}& \cc{65.62}& \cc{26.38}& \cc{13.48}& \cc{27.09}& \cc{25.75}& \cc{23.88}& \cc{28.21}& \cc{24.38}& \cc{28.34}& \cc{34.75}& \cc{11.25}& \cc{28.75}& \cc{27.75}& \cc{31.38}& \cc{25.5}& \cc{26.12}& \cc{21.12}& \\
 heb\_Hebr& Hebrew& \cc{33.83}& \cc{86.5}& \cc{59.88}& \cc{25.34}& \cc{31.21}& \cc{21.88}& \cc{19.38}& \cc{31.84}& \cc{25.5}& \cc{30.34}& \cc{41.5}& \cc{34.88}& \cc{41.5}& \cc{36.62}& \cc{35.38}& \cc{30.5}& \cc{64.88}& \cc{70.5}& \\
 henan& henan& \cc{83.77}& \cc{98}& \cc{45.88}& \cc{14.98}& \cc{82.52}& \cc{26.75}& \cc{20.75}& \cc{85.52}& \cc{26.25}& \cc{69.04}& \cc{45.12}& \cc{28.38}& \cc{80.88}& \cc{79.25}& \cc{64}& \cc{53.12}& \cc{58.5}& \cc{54.88}& \\
 hin\_Deva& Hindi& \cc{57.68}& \cc{70.75}& \cc{43.38}& \cc{22.72}& \cc{45.82}& \cc{17.5}& \cc{30.5}& \cc{60.92}& \cc{25.88}& \cc{31.96}& \cc{32}& \cc{29.62}& \cc{62.5}& \cc{67.75}& \cc{53.25}& \cc{43.75}& \cc{49.75}& \cc{38.5}& \\
 hubei& hubei& \cc{84.14}& \cc{90.38}& \cc{75}& \cc{20.35}& \cc{83.27}& \cc{24.5}& \cc{22}& \cc{85.27}& \cc{26.12}& \cc{71.29}& \cc{35.25}& \cc{45.12}& \cc{81}& \cc{78}& \cc{68.12}& \cc{53.75}& \cc{61.5}& \cc{59.12}& \\
 hun\_Latn& Hungarian& \cc{32.08}& \cc{94.88}& \cc{76.88}& \cc{23.35}& \cc{30.34}& \cc{18.75}& \cc{3.38}& \cc{31.71}& \cc{26.88}& \cc{27.72}& \cc{53.25}& \cc{55.62}& \cc{56.12}& \cc{65.5}& \cc{55.38}& \cc{53.62}& \cc{60.5}& \cc{71.62}& \\
 hunan& hunan& \cc{78.53}& \cc{76.62}& \cc{56.12}& \cc{21.22}& \cc{79.03}& \cc{13.75}& \cc{16.38}& \cc{81.02}& \cc{24.38}& \cc{64.92}& \cc{59.5}& \cc{60.12}& \cc{74.25}& \cc{71.38}& \cc{59.88}& \cc{50.12}& \cc{59.62}& \cc{68.62}& \\
 hye\_Armn& Armenian& \cc{29.21}& \cc{79.75}& \cc{53.25}& \cc{23.72}& \cc{28.34}& \cc{23.5}& \cc{31.38}& \cc{31.09}& \cc{25}& \cc{30.59}& \cc{31.88}& \cc{46.88}& \cc{24.25}& \cc{21.12}& \cc{21.38}& \cc{21.88}& \cc{33.12}& \cc{34.88}& \\
 ibo\_Latn& Igbo& \cc{29.21}& \cc{45}& \cc{23.12}& \cc{11.49}& \cc{25.97}& \cc{25.38}& \cc{21.38}& \cc{26.59}& \cc{25.62}& \cc{29.96}& \cc{16.38}& \cc{30.62}& \cc{28.62}& \cc{26.62}& \cc{21.88}& \cc{27.25}& \cc{24.75}& \cc{9}& \\
 ind\_Latn& Indonesian& \cc{78.03}& \cc{95.12}& \cc{69.5}& \cc{23.85}& \cc{41.7}& \cc{28}& \cc{19.12}& \cc{79.65}& \cc{33.12}& \cc{36.08}& \cc{68.5}& \cc{64.5}& \cc{82.25}& \cc{82.38}& \cc{66.25}& \cc{60}& \cc{68.12}& \cc{81.88}& \\
 isl\_Latn& Icelandic& \cc{37.7}& \cc{98}& \cc{61.12}& \cc{22.6}& \cc{31.21}& \cc{22.75}& \cc{26.38}& \cc{33.21}& \cc{24.12}& \cc{30.59}& \cc{36.25}& \cc{18.75}& \cc{27.88}& \cc{26}& \cc{18.5}& \cc{22.25}& \cc{23.88}& \cc{43.88}& \\
 ita\_Latn& Italian& \cc{85.89}& \cc{96}& \cc{73.5}& \cc{24.72}& \cc{57.43}& \cc{16.12}& \cc{24.12}& \cc{85.02}& \cc{24.25}& \cc{54.18}& \cc{64.88}& \cc{56.12}& \cc{86.25}& \cc{87.25}& \cc{67.5}& \cc{55}& \cc{91.38}& \cc{79.12}& \\
 jav\_Latn& Javanese& \cc{42.95}& \cc{95}& \cc{51.12}& \cc{19.98}& \cc{30.96}& \cc{19.75}& \cc{15}& \cc{41.45}& \cc{24.12}& \cc{31.34}& \cc{25.62}& \cc{15.38}& \cc{29.62}& \cc{35.5}& \cc{35.38}& \cc{26.25}& \cc{55.75}& \cc{13.75}& \\
 jiangxi& jiangxi& \cc{81.77}& \cc{93.38}& \cc{50.5}& \cc{23.35}& \cc{81.27}& \cc{18.38}& \cc{15}& \cc{83.27}& \cc{25.62}& \cc{68.29}& \cc{55.62}& \cc{58.88}& \cc{72}& \cc{75}& \cc{56.5}& \cc{47.62}& \cc{64}& \cc{65.62}& \\
 jpn\_Jpan& Japanese& \cc{73.28}& \cc{83}& \cc{81}& \cc{24.97}& \cc{33.21}& \cc{24.38}& \cc{16.75}& \cc{75.41}& \cc{24.88}& \cc{59.93}& \cc{66.38}& \cc{48.75}& \cc{74.5}& \cc{75}& \cc{54.5}& \cc{49.62}& \cc{89}& \cc{69}& \\
 kan\_Knda& Kannada& \cc{33.71}& \cc{88.5}& \cc{38.62}& \cc{23.35}& \cc{35.71}& \cc{16}& \cc{13.75}& \cc{33.33}& \cc{24.12}& \cc{30.59}& \cc{38.25}& \cc{20.38}& \cc{23}& \cc{24.12}& \cc{22.88}& \cc{18.75}& \cc{41.88}& \cc{23.88}& \\
 kat\_Geor& Georgian& \cc{30.71}& \cc{85.12}& \cc{56.25}& \cc{21.47}& \cc{27.72}& \cc{23.88}& \cc{18.38}& \cc{30.84}& \cc{24.38}& \cc{29.59}& \cc{44}& \cc{33.38}& \cc{29.25}& \cc{26.25}& \cc{26.62}& \cc{28}& \cc{28}& \cc{26.38}& \\
 kaz\_Cyrl& Kazakh& \cc{33.21}& \cc{85.25}& \cc{53.62}& \cc{23.97}& \cc{32.58}& \cc{29.75}& \cc{45}& \cc{36.2}& \cc{25.25}& \cc{28.96}& \cc{50.75}& \cc{45.12}& \cc{30.12}& \cc{21.88}& \cc{28.62}& \cc{22.5}& \cc{35}& \cc{33.25}& \\
 kea\_Latn& Kabuverdianu& \cc{52.31}& \cc{88.62}& \cc{34}& \cc{21.22}& \cc{40.32}& \cc{23.12}& \cc{13.88}& \cc{50.81}& \cc{24}& \cc{35.71}& \cc{41.25}& \cc{45.62}& \cc{37.88}& \cc{35.88}& \cc{31.25}& \cc{27.5}& \cc{38.12}& \cc{38.38}& \\
 khk\_Cyrl& Halh Mongolian& \cc{30.71}& \cc{75.12}& \cc{18.5}& \cc{19.85}& \cc{28.21}& \cc{16.62}& \cc{23.5}& \cc{26.72}& \cc{24.62}& \cc{28.34}& \cc{25.5}& \cc{20}& \cc{29.38}& \cc{22.62}& \cc{29.25}& \cc{26.75}& \cc{19.88}& \cc{11}& \\
 khm\_Khmr& Khmer& \cc{31.09}& \cc{81.62}& \cc{55.25}& \cc{21.6}& \cc{28.59}& \cc{29.38}& \cc{4.38}& \cc{28.21}& \cc{27.38}& \cc{31.09}& \cc{35.88}& \cc{11.88}& \cc{24}& \cc{23.25}& \cc{23.75}& \cc{17.75}& \cc{25.88}& \cc{20.75}& \\
 kor\_Hang& Korean& \cc{76.15}& \cc{90.25}& \cc{73.88}& \cc{22.1}& \cc{35.58}& \cc{28.88}& \cc{28.38}& \cc{79.78}& \cc{28.38}& \cc{48.81}& \cc{64.12}& \cc{58.62}& \cc{80.12}& \cc{79.25}& \cc{72.5}& \cc{55.62}& \cc{71.38}& \cc{78.38}& \\
 lao\_Laoo& Lao& \cc{36.33}& \cc{81.5}& \cc{29.88}& \cc{23.72}& \cc{27.84}& \cc{19.5}& \cc{13.25}& \cc{36.95}& \cc{23.88}& \cc{30.34}& \cc{39.75}& \cc{16.5}& \cc{32.5}& \cc{32.38}& \cc{30}& \cc{30.88}& \cc{46.25}& \cc{20.12}& \\
 lit\_Latn& Lithuanian& \cc{38.2}& \cc{86}& \cc{73.12}& \cc{23.72}& \cc{34.33}& \cc{16.25}& \cc{24.75}& \cc{38.83}& \cc{26.12}& \cc{30.21}& \cc{63}& \cc{41.25}& \cc{27.12}& \cc{28.25}& \cc{30.5}& \cc{30}& \cc{63}& \cc{80.38}& \\
 lug\_Latn& Ganda& \cc{29.59}& \cc{59.25}& \cc{19.88}& \cc{13.48}& \cc{28.84}& \cc{27.62}& \cc{30}& \cc{26.22}& \cc{32.25}& \cc{27.22}& \cc{24.5}& \cc{20.38}& \cc{18.62}& \cc{15.25}& \cc{28.38}& \cc{23.88}& \cc{24.88}& \cc{35}& \\
 luo\_Latn& Luo& \cc{31.59}& \cc{36.75}& \cc{30.12}& \cc{16.48}& \cc{26.47}& \cc{21.75}& \cc{15.12}& \cc{27.22}& \cc{25}& \cc{27.72}& \cc{33.5}& \cc{35.62}& \cc{25.12}& \cc{22.88}& \cc{24.12}& \cc{24.38}& \cc{44}& \cc{14}& \\
 lvs\_Latn& Standard Latvian& \cc{35.58}& \cc{98}& \cc{71.12}& \cc{24.72}& \cc{31.09}& \cc{20.75}& \cc{8.5}& \cc{33.58}& \cc{26}& \cc{29.21}& \cc{44.38}& \cc{26.75}& \cc{24.75}& \cc{19.88}& \cc{24.88}& \cc{36.5}& \cc{61.25}& \cc{64.5}& \\
 mal\_Mlym& Malayalam& \cc{34.46}& \cc{80.12}& \cc{29.75}& \cc{23.47}& \cc{32.71}& \cc{18.62}& \cc{31}& \cc{33.08}& \cc{23.88}& \cc{29.59}& \cc{16.38}& \cc{28.38}& \cc{26.12}& \cc{22.5}& \cc{27.5}& \cc{22.88}& \cc{30.38}& \cc{12}& \\
 mar\_Deva& Marathi& \cc{43.82}& \cc{85.5}& \cc{55.5}& \cc{23.47}& \cc{41.57}& \cc{14.75}& \cc{18.62}& \cc{40.95}& \cc{29.62}& \cc{31.46}& \cc{39.12}& \cc{35.5}& \cc{46.25}& \cc{41.12}& \cc{39.25}& \cc{27.75}& \cc{34.5}& \cc{43.38}& \\
 minnan& minnan& \cc{45.82}& \cc{55.25}& \cc{33.38}& \cc{21.97}& \cc{36.83}& \cc{23.88}& \cc{25.5}& \cc{36.95}& \cc{26.12}& \cc{48.81}& \cc{36.5}& \cc{40.12}& \cc{52.75}& \cc{48.75}& \cc{37.62}& \cc{32.38}& \cc{44.88}& \cc{36.38}& \\
 mkd\_Cyrl& Macedonian& \cc{51.81}& \cc{78.62}& \cc{66}& \cc{24.34}& \cc{34.83}& \cc{12.62}& \cc{16.75}& \cc{45.07}& \cc{29.88}& \cc{30.21}& \cc{51.5}& \cc{37}& \cc{56.88}& \cc{68.5}& \cc{52.62}& \cc{37.88}& \cc{66}& \cc{64.62}& \\
 mlt\_Latn& Maltese& \cc{40.7}& \cc{83.5}& \cc{16}& \cc{23.47}& \cc{35.33}& \cc{27.88}& \cc{25.88}& \cc{36.08}& \cc{24.12}& \cc{31.34}& \cc{28}& \cc{41.25}& \cc{30.88}& \cc{21.88}& \cc{30.38}& \cc{28.25}& \cc{25.12}& \cc{25.25}& \\
 mya\_Mymr& Burmese& \cc{29.96}& \cc{78}& \cc{26}& \cc{16.35}& \cc{28.21}& \cc{26.88}& \cc{28}& \cc{28.96}& \cc{25.38}& \cc{29.21}& \cc{10.25}& \cc{39.25}& \cc{24.5}& \cc{27.38}& \cc{22.25}& \cc{30.88}& \cc{32}& \cc{28.38}& \\
 ningxia& ningxia& \cc{71.41}& \cc{79.5}& \cc{33.5}& \cc{21.35}& \cc{72.28}& \cc{23.25}& \cc{21.25}& \cc{75.78}& \cc{25.88}& \cc{67.67}& \cc{31.5}& \cc{19.88}& \cc{67.62}& \cc{71.12}& \cc{50.5}& \cc{41.25}& \cc{56.5}& \cc{35.62}& \\
 nld\_Latn& Dutch& \cc{76.78}& \cc{94.38}& \cc{78.5}& \cc{24.22}& \cc{43.7}& \cc{16.62}& \cc{29.25}& \cc{80.65}& \cc{31.38}& \cc{39.95}& \cc{53.75}& \cc{51.38}& \cc{76}& \cc{73.5}& \cc{67.38}& \cc{51.25}& \cc{68.12}& \cc{61.25}& \\
 nob\_Latn& Norwegian Bokmål& \cc{49.69}& \cc{84}& \cc{58.88}& \cc{19.98}& \cc{35.96}& \cc{25}& \cc{20.12}& \cc{33.21}& \cc{25.12}& \cc{30.21}& \cc{55.38}& \cc{38.75}& \cc{76.25}& \cc{75.12}& \cc{55}& \cc{33.38}& \cc{73}& \cc{68.12}& \\
 npi\_Deva& Nepali& \cc{33.21}& \cc{98}& \cc{70.12}& \cc{24.84}& \cc{32.58}& \cc{16.62}& \cc{10.62}& \cc{30.59}& \cc{28.62}& \cc{31.96}& \cc{28}& \cc{31}& \cc{37}& \cc{36}& \cc{38.38}& \cc{28.12}& \cc{33.88}& \cc{35.12}& \\
 ory\_Orya& Odia& \cc{40.57}& \cc{86.38}& \cc{49.38}& \cc{22.35}& \cc{33.21}& \cc{15}& \cc{26.25}& \cc{34.46}& \cc{24}& \cc{31.09}& \cc{26.75}& \cc{41.38}& \cc{22.88}& \cc{23.5}& \cc{22.88}& \cc{31.62}& \cc{36.5}& \cc{18.12}& \\
 pan\_Guru& Eastern Panjabi& \cc{43.45}& \cc{80.12}& \cc{45.88}& \cc{23.85}& \cc{34.71}& \cc{29.25}& \cc{20.38}& \cc{41.7}& \cc{26}& \cc{27.09}& \cc{31.75}& \cc{28.62}& \cc{42.25}& \cc{42}& \cc{32.88}& \cc{28.25}& \cc{31.25}& \cc{31.25}& \\
 pbt\_Arab& Southern Pashto& \cc{34.08}& \cc{69.5}& \cc{44}& \cc{20.6}& \cc{29.59}& \cc{34}& \cc{13.25}& \cc{31.46}& \cc{24.5}& \cc{29.84}& \cc{46.12}& \cc{20.75}& \cc{20.25}& \cc{19.62}& \cc{26.75}& \cc{20.5}& \cc{16.5}& \cc{24.88}& \\
 pes\_Arab& Western Persian& \cc{36.45}& \cc{95}& \cc{59.38}& \cc{23.47}& \cc{33.46}& \cc{34}& \cc{19.88}& \cc{34.08}& \cc{22.12}& \cc{29.96}& \cc{43.38}& \cc{35.88}& \cc{62.88}& \cc{57.75}& \cc{42.75}& \cc{35.38}& \cc{68.12}& \cc{60.62}& \\
 pol\_Latn& Polish& \cc{65.29}& \cc{94.62}& \cc{84.38}& \cc{24.09}& \cc{34.71}& \cc{21}& \cc{18.5}& \cc{42.45}& \cc{24.62}& \cc{31.21}& \cc{68.5}& \cc{44.38}& \cc{79.75}& \cc{81.38}& \cc{62.5}& \cc{51.38}& \cc{80.62}& \cc{81.62}& \\
 por\_Latn& Portuguese& \cc{86.27}& \cc{98}& \cc{85.38}& \cc{22.22}& \cc{52.81}& \cc{20.88}& \cc{14.62}& \cc{85.39}& \cc{25.5}& \cc{53.93}& \cc{56.62}& \cc{49.12}& \cc{79.5}& \cc{83.25}& \cc{62.5}& \cc{50.25}& \cc{78.75}& \cc{79.5}& \\
 ron\_Latn& Romanian& \cc{52.43}& \cc{98}& \cc{89.88}& \cc{24.84}& \cc{37.2}& \cc{9}& \cc{16}& \cc{43.57}& \cc{28.88}& \cc{34.08}& \cc{64.5}& \cc{59.25}& \cc{65.62}& \cc{67.38}& \cc{62.88}& \cc{52.25}& \cc{80.75}& \cc{94.38}& \\
 rus\_Cyrl& Russian& \cc{85.64}& \cc{93.38}& \cc{84.88}& \cc{24.59}& \cc{46.69}& \cc{6.62}& \cc{24.5}& \cc{84.39}& \cc{33.88}& \cc{44.44}& \cc{59.38}& \cc{53.75}& \cc{85.38}& \cc{85.88}& \cc{69.88}& \cc{56.25}& \cc{83.25}& \cc{83.62}& \\
 shan1xi& shan1xi& \cc{72.91}& \cc{78.25}& \cc{59.88}& \cc{17.85}& \cc{73.66}& \cc{17}& \cc{26.25}& \cc{78.78}& \cc{24}& \cc{65.79}& \cc{56.75}& \cc{58.88}& \cc{69}& \cc{72.12}& \cc{55.88}& \cc{49.62}& \cc{59.88}& \cc{76.62}& \\
 shan3xi& shan3xi& \cc{77.65}& \cc{78.88}& \cc{49}& \cc{18.98}& \cc{76.65}& \cc{20.75}& \cc{23}& \cc{79.65}& \cc{25.12}& \cc{74.03}& \cc{44.12}& \cc{34.38}& \cc{75.75}& \cc{72.12}& \cc{60.5}& \cc{57.5}& \cc{50.88}& \cc{49.62}& \\
 shandong& shandong& \cc{79.28}& \cc{78.5}& \cc{60}& \cc{16.73}& \cc{84.02}& \cc{28.12}& \cc{18.38}& \cc{82.15}& \cc{24.12}& \cc{70.79}& \cc{46.25}& \cc{45.38}& \cc{77.12}& \cc{80.88}& \cc{54.38}& \cc{51}& \cc{75.12}& \cc{75.25}& \\
 shanghai& shanghai& \cc{63.92}& \cc{63.75}& \cc{30.12}& \cc{22.97}& \cc{54.81}& \cc{25.12}& \cc{15.88}& \cc{66.42}& \cc{25.38}& \cc{57.18}& \cc{41.5}& \cc{38}& \cc{54.5}& \cc{55.62}& \cc{38}& \cc{41.12}& \cc{29.75}& \cc{13.62}& \\
 sichuan& sichuan& \cc{80.15}& \cc{75.88}& \cc{48.62}& \cc{23.47}& \cc{83.27}& \cc{23.62}& \cc{20.88}& \cc{82.27}& \cc{26}& \cc{70.54}& \cc{48}& \cc{44.25}& \cc{71.5}& \cc{73.75}& \cc{56.12}& \cc{46.38}& \cc{59.38}& \cc{54.12}& \\
 sin\_Sinh& Sinhala& \cc{34.08}& \cc{60}& \cc{30.88}& \cc{23.35}& \cc{33.58}& \cc{23.38}& \cc{14.38}& \cc{34.46}& \cc{25.38}& \cc{30.09}& \cc{14.38}& \cc{36.25}& \cc{27.5}& \cc{24.62}& \cc{24.25}& \cc{19.62}& \cc{26.62}& \cc{28.25}& \\
 slk\_Latn& Slovak& \cc{56.93}& \cc{96.75}& \cc{84}& \cc{22.97}& \cc{38.83}& \cc{10.25}& \cc{18.5}& \cc{40.95}& \cc{24.5}& \cc{30.21}& \cc{69}& \cc{46.62}& \cc{54.62}& \cc{60.25}& \cc{57.5}& \cc{45.12}& \cc{73.38}& \cc{79.75}& \\
 slv\_Latn& Slovenian& \cc{45.32}& \cc{98}& \cc{74.5}& \cc{24.22}& \cc{37.08}& \cc{17.75}& \cc{14.38}& \cc{37.33}& \cc{23.62}& \cc{29.59}& \cc{59.12}& \cc{50.88}& \cc{29.25}& \cc{32.5}& \cc{34.25}& \cc{28.38}& \cc{68.5}& \cc{64.62}& \\
 sna\_Latn& Shona& \cc{31.59}& \cc{61.12}& \cc{3.75}& \cc{16.48}& \cc{27.09}& \cc{24.75}& \cc{21.38}& \cc{33.83}& \cc{24}& \cc{29.21}& \cc{9.88}& \cc{5.38}& \cc{26.5}& \cc{21}& \cc{28}& \cc{27.88}& \cc{30.38}& \cc{13.88}& \\
 snd\_Arab& Sindhi& \cc{36.7}& \cc{81.38}& \cc{24.88}& \cc{20.72}& \cc{31.59}& \cc{30.5}& \cc{20.25}& \cc{34.46}& \cc{25.62}& \cc{31.96}& \cc{14.5}& \cc{18.25}& \cc{25.62}& \cc{23.5}& \cc{23.75}& \cc{20.88}& \cc{40}& \cc{2}& \\
 som\_Latn& Somali& \cc{28.71}& \cc{73}& \cc{15.12}& \cc{23.35}& \cc{26.84}& \cc{22.88}& \cc{8.25}& \cc{30.34}& \cc{24.88}& \cc{28.84}& \cc{16.12}& \cc{30.88}& \cc{27.12}& \cc{25}& \cc{24.88}& \cc{37.75}& \cc{37}& \cc{25.5}& \\
 spa\_Latn& Spanish& \cc{79.15}& \cc{95.5}& \cc{70.38}& \cc{21.22}& \cc{55.06}& \cc{20.88}& \cc{36.62}& \cc{79.53}& \cc{24.38}& \cc{53.93}& \cc{60.5}& \cc{49.25}& \cc{76.38}& \cc{79.88}& \cc{62.5}& \cc{56.12}& \cc{89.25}& \cc{90.88}& \\
 srp\_Cyrl& Serbian& \cc{54.43}& \cc{96}& \cc{82}& \cc{24.84}& \cc{40.32}& \cc{15.38}& \cc{31.5}& \cc{43.82}& \cc{25.62}& \cc{31.21}& \cc{58.12}& \cc{48.12}& \cc{55.5}& \cc{70.62}& \cc{49.38}& \cc{44.75}& \cc{83.38}& \cc{85.25}& \\
 suhang& suhang& \cc{65.29}& \cc{65.38}& \cc{38.62}& \cc{22.22}& \cc{60.92}& \cc{29.88}& \cc{20.62}& \cc{71.54}& \cc{26}& \cc{59.18}& \cc{19}& \cc{30.5}& \cc{56.38}& \cc{58.62}& \cc{49.25}& \cc{43.75}& \cc{23.5}& \cc{35}& \\
 sun\_Latn& Sundanese& \cc{39.58}& \cc{79.25}& \cc{31.38}& \cc{24.09}& \cc{34.33}& \cc{28.62}& \cc{28.38}& \cc{39.08}& \cc{23}& \cc{30.21}& \cc{24.12}& \cc{23}& \cc{32.88}& \cc{33.62}& \cc{37.5}& \cc{28.38}& \cc{38.88}& \cc{24.38}& \\
 swe\_Latn& Swedish& \cc{57.43}& \cc{95.88}& \cc{71.75}& \cc{23.72}& \cc{44.07}& \cc{23.62}& \cc{16.88}& \cc{37.95}& \cc{24}& \cc{32.46}& \cc{64.12}& \cc{53.5}& \cc{75.88}& \cc{79.5}& \cc{61}& \cc{48.38}& \cc{84.75}& \cc{83.88}& \\
 swh\_Latn& Swahili& \cc{34.33}& \cc{76}& \cc{26.75}& \cc{19.1}& \cc{30.46}& \cc{19.5}& \cc{29.12}& \cc{32.58}& \cc{25.25}& \cc{26.09}& \cc{31.5}& \cc{30.12}& \cc{33.5}& \cc{36}& \cc{38.75}& \cc{27.5}& \cc{31.88}& \cc{9}& \\
 tam\_Taml& Tamil& \cc{33.83}& \cc{80.38}& \cc{38.12}& \cc{22.47}& \cc{34.08}& \cc{25.75}& \cc{17}& \cc{32.21}& \cc{25.38}& \cc{27.59}& \cc{38.25}& \cc{39}& \cc{33.5}& \cc{29.88}& \cc{28.75}& \cc{23.12}& \cc{40.88}& \cc{36.62}& \\
 tel\_Telu& Telugu& \cc{34.58}& \cc{88.5}& \cc{40.5}& \cc{22.72}& \cc{33.71}& \cc{22}& \cc{19.38}& \cc{32.83}& \cc{25.25}& \cc{29.21}& \cc{19.5}& \cc{36.12}& \cc{27.38}& \cc{27.62}& \cc{30}& \cc{21.62}& \cc{16.5}& \cc{29.25}& \\
 tgk\_Cyrl& Tajik& \cc{36.58}& \cc{80.5}& \cc{35.12}& \cc{23.22}& \cc{27.84}& \cc{36.5}& \cc{10.38}& \cc{34.83}& \cc{24.25}& \cc{29.21}& \cc{18.88}& \cc{25.62}& \cc{38.12}& \cc{31.25}& \cc{39.25}& \cc{26.62}& \cc{28.62}& \cc{15.75}& \\
 tgl\_Latn& Tagalog& \cc{53.43}& \cc{95.62}& \cc{83.75}& \cc{22.97}& \cc{47.07}& \cc{21.88}& \cc{30}& \cc{41.32}& \cc{25}& \cc{30.34}& \cc{51.38}& \cc{24.62}& \cc{47.75}& \cc{59.25}& \cc{42.88}& \cc{31.75}& \cc{51.75}& \cc{48.5}& \\
 tha\_Thai& Thai& \cc{71.04}& \cc{90.12}& \cc{78}& \cc{25.09}& \cc{36.33}& \cc{22.12}& \cc{15.5}& \cc{69.16}& \cc{23.88}& \cc{29.21}& \cc{58.38}& \cc{50.38}& \cc{81.38}& \cc{74.88}& \cc{62.12}& \cc{48.38}& \cc{93.88}& \cc{75.38}& \\
 tianjin& tianjin& \cc{84.89}& \cc{98}& \cc{73.62}& \cc{15.48}& \cc{86.77}& \cc{21.75}& \cc{15.12}& \cc{87.02}& \cc{26.12}& \cc{72.53}& \cc{63.62}& \cc{71.5}& \cc{77.75}& \cc{79.12}& \cc{71.62}& \cc{62.38}& \cc{73.62}& \cc{81.5}& \\
 tur\_Latn& Turkish& \cc{54.43}& \cc{89.62}& \cc{73}& \cc{23.97}& \cc{36.95}& \cc{22.88}& \cc{24.62}& \cc{42.07}& \cc{25.75}& \cc{29.46}& \cc{65.38}& \cc{40.5}& \cc{70.62}& \cc{77.62}& \cc{54.88}& \cc{50.25}& \cc{75.38}& \cc{84.88}& \\
 ukr\_Cyrl& Ukrainian& \cc{55.68}& \cc{96.62}& \cc{81.75}& \cc{24.47}& \cc{38.83}& \cc{22.38}& \cc{36.75}& \cc{48.56}& \cc{24.38}& \cc{32.21}& \cc{58.12}& \cc{48.12}& \cc{59.12}& \cc{70}& \cc{53.75}& \cc{44.38}& \cc{64}& \cc{81.12}& \\
 urd\_Arab& Urdu& \cc{55.68}& \cc{93.88}& \cc{74}& \cc{24.09}& \cc{47.19}& \cc{18.25}& \cc{13.75}& \cc{56.3}& \cc{31.12}& \cc{31.84}& \cc{26.38}& \cc{39.25}& \cc{49.62}& \cc{63.5}& \cc{54.75}& \cc{40.5}& \cc{31.38}& \cc{45.88}& \\
 uzn\_Latn& Northern Uzbek& \cc{37.83}& \cc{98}& \cc{69.12}& \cc{24.09}& \cc{36.83}& \cc{17.38}& \cc{25.5}& \cc{36.2}& \cc{30.38}& \cc{29.34}& \cc{25}& \cc{34.75}& \cc{33.75}& \cc{29.88}& \cc{28.75}& \cc{24.88}& \cc{36.12}& \cc{8.38}& \\
 vie\_Latn& Vietnamese& \cc{75.91}& \cc{89.62}& \cc{64.25}& \cc{23.35}& \cc{28.46}& \cc{21.25}& \cc{21.38}& \cc{77.03}& \cc{28.25}& \cc{29.71}& \cc{51.5}& \cc{53.75}& \cc{73.88}& \cc{66.38}& \cc{60.88}& \cc{55.88}& \cc{73.38}& \cc{60.25}& \\
 wol\_Latn& Wolof& \cc{29.34}& \cc{74}& \cc{25.25}& \cc{15.61}& \cc{26.84}& \cc{22.5}& \cc{26.25}& \cc{26.84}& \cc{24.5}& \cc{30.46}& \cc{19.38}& \cc{16.25}& \cc{22.5}& \cc{20.38}& \cc{27.62}& \cc{17.25}& \cc{19.88}& \cc{30.75}& \\
 wuzhong& wuzhong& \cc{85.27}& \cc{95.38}& \cc{76.5}& \cc{19.73}& \cc{85.89}& \cc{19.88}& \cc{17}& \cc{86.39}& \cc{25.88}& \cc{74.16}& \cc{65.25}& \cc{58.38}& \cc{75.5}& \cc{81.62}& \cc{66.5}& \cc{52}& \cc{79.5}& \cc{84.12}& \\
 xho\_Latn& Xhosa& \cc{28.46}& \cc{65.62}& \cc{17}& \cc{14.86}& \cc{26.97}& \cc{19.5}& \cc{29.62}& \cc{29.21}& \cc{25.38}& \cc{29.84}& \cc{30.62}& \cc{26.5}& \cc{22.5}& \cc{23.62}& \cc{30.75}& \cc{27.75}& \cc{8.25}& \cc{3.62}& \\
 yor\_Latn& Yoruba& \cc{29.84}& \cc{59.12}& \cc{2}& \cc{12.86}& \cc{28.34}& \cc{34.25}& \cc{23.12}& \cc{29.46}& \cc{25.62}& \cc{29.21}& \cc{41.25}& \cc{13.38}& \cc{23.88}& \cc{26.25}& \cc{30.25}& \cc{19.12}& \cc{13.25}& \cc{25.88}& \\
 yunnan& yunnan& \cc{78.4}& \cc{90.5}& \cc{56}& \cc{23.85}& \cc{78.4}& \cc{22.75}& \cc{10.5}& \cc{80.52}& \cc{24.38}& \cc{68.54}& \cc{50.38}& \cc{38.25}& \cc{74}& \cc{80}& \cc{55.75}& \cc{50.62}& \cc{70.12}& \cc{61}& \\
 zho\_Hans& Chinese (Simplified)& \cc{87.02}& \cc{98}& \cc{74.12}& \cc{22.47}& \cc{87.02}& \cc{23}& \cc{18.5}& \cc{87.39}& \cc{25.88}& \cc{70.04}& \cc{65.38}& \cc{71.12}& \cc{86.62}& \cc{86.88}& \cc{68}& \cc{57.5}& \cc{81.5}& \cc{83.25}& \\
 zho\_Hant& Chinese (Traditional)& \cc{87.39}& \cc{88.88}& \cc{98}& \cc{15.98}& \cc{85.52}& \cc{19.12}& \cc{26.38}& \cc{86.89}& \cc{31}& \cc{74.91}& \cc{68.12}& \cc{59.88}& \cc{79}& \cc{83}& \cc{74.12}& \cc{56.75}& \cc{61.12}& \cc{63.5}& \\
 zsm\_Latn& Standard Malay& \cc{71.04}& \cc{98}& \cc{74.5}& \cc{23.6}& \cc{55.68}& \cc{14.62}& \cc{20}& \cc{66.67}& \cc{25.75}& \cc{32.33}& \cc{48.38}& \cc{44.25}& \cc{78.75}& \cc{84.25}& \cc{62.38}& \cc{50.88}& \cc{75.5}& \cc{85.75}& \\
 zul\_Latn& Zulu& \cc{30.84}& \cc{56.75}& \cc{35.75}& \cc{21.97}& \cc{29.84}& \cc{28.75}& \cc{28}& \cc{29.09}& \cc{25.88}& \cc{30.59}& \cc{21}& \cc{31}& \cc{16.62}& \cc{21.12}& \cc{30.75}& \cc{19.75}& \cc{19.62}& \cc{9}& \\
 \midrule
\multicolumn{2}{c}{Average} & \cc{52.88}& \cc{85.30}& \cc{56.63}& \cc{21.88}& \cc{43.51}& \cc{21.83}& \cc{20.96}& \cc{50.89}& \cc{25.94}& \cc{40.52}& \cc{43.59}& \cc{39.01}& \cc{52.29}& \cc{52.66}& \cc{45.00}& \cc{38.21}& \cc{53.86}& \cc{51.56}& \\
 \bottomrule
\end{tabular}%
}
\end{table*}

\end{document}